\documentclass[lettersize,journal]{IEEEtran}
\usepackage{amsmath,amsfonts}
\usepackage{algorithm}
\usepackage{array}
\usepackage[caption=false,font=normalsize,labelfont=sf,textfont=sf]{subfig}
\usepackage{textcomp}
\usepackage{stfloats}
\usepackage{url}
\usepackage{verbatim}
\usepackage{graphicx}
\usepackage{cite}
\usepackage{booktabs}
\usepackage[pagebackref=true,breaklinks=true,letterpaper=true,colorlinks,citecolor=blue,linkcolor=blue,bookmarks=false]{hyperref}
\usepackage{xcolor}
\usepackage{algpseudocode}
 \usepackage{multirow}
 
\begin{document}

\title{RAGDiffusion++: From Macro-Retrieval to Micro-Fidelity Alignment for Garment Generation}

\author{
Yuhan Li$^{*}$, 
Xianfeng Tan$^{*}$,
Fangao Zeng,
Wenxiang Shang,
Pipei Huang,
Hao Zhou,
Zhiyu Jin,
Wenjun Zhang,
Bingbing Ni
\thanks{
\indent Yuhan Li, Wenjun Zhang and Bingbing Ni are with Shanghai Jiao Tong University, Shanghai, P.R.China.\\
\indent Xianfeng Tan, Fangao Zeng, Wenxiang Shang, Pipei Huang, Hao Zhou, and Zhiyu Jin are with Alibaba Group, P.R.China.\\
\indent $^*$Equal Contribution. \quad Corresponding author: Bingbing Ni.}
}

\markboth{Submitted to IEEE Transactions on Pattern Analysis and Machine Intelligence,~Vol.~XX, No.~XX, XX~2026}%
{Li \MakeLowercase{\textit{et al.}}: RAGDiffusion++: From Macro-Retrieval to Micro-Fidelity Alignment for Garment Generation}



\maketitle

\begin{abstract}
    Standard clothing asset generation---restoring forward-facing flat-lay garment images from diverse real-world contexts---holds immense commercial value yet demands both macroscopic topological accuracy and microscopic physical fidelity. Although our previous work RAGDiffusion effectively eradicated large-scale structural hallucinations via retrieval-augmented macro-constraints, achieving industrial-grade micro-texture realism remains an unsolved bottleneck. We formally identify this limitation as \textit{High-Frequency Trajectory Collapse}: supervised fine-tuning (SFT) converges to the conditional mean of the training distribution, which is dominated by smooth, low-frequency textures, causing high-frequency patterns (e.g., fabric weaves, intricate logos) to become nearly un-sampleable. Na\"{\i}vely applying Reinforcement Learning (RL) post-training further triggers \textit{Artifact Hacking}, where models exploit semantic biases in generic reward models by generating deceptive checkerboard noise.
    Our key insight is that RL can fundamentally reshape the sampling distribution of flow models---elevating the probability of high-fidelity trajectories under accurate reward guidance---while adversarial regularization prevents exploitation of reward blind spots. Realizing this principle requires three prerequisites: (i)~\textit{inherent capacity}, established through a 27,725-pair high-complexity garment dataset (STGarment-Plus) and a Dual-Image-Stream FLUX architecture upgrade; (ii)~\textit{perceptive reward}, provided by a novel attribute-aware reward model (Garment-RM) trained on 500K images via fine-grained contrastive learning, achieving 84.67\% human preference accuracy; and (iii)~\textit{hacking prevention}, enforced by our Adversarial-Regularized GRPO (AR-GRPO) strategy that integrates a dynamic discriminator into the RL sampling trajectory to penalize artifacts while enriching authentic high-frequency details.
    Extensive experiments demonstrate that RAGDiffusion++ reduces FID by 14.3\% and KID by 34.2\% over our preliminary RAGDiffusion, surpasses the best open-source baseline by 51.0\% in KID, and outperforms the large-scale commercial model Nano Banana by 10.9\% in FID, setting a new benchmark for commercial-grade garment synthesis. We will open-source the STGarment-Plus dataset and the pre-trained Garment-RM to facilitate future research.
    \end{abstract}
    

\section{Introduction}
\label{sec:intro}

\begin{figure}[t]
  \centering
  \includegraphics[width=1\linewidth]{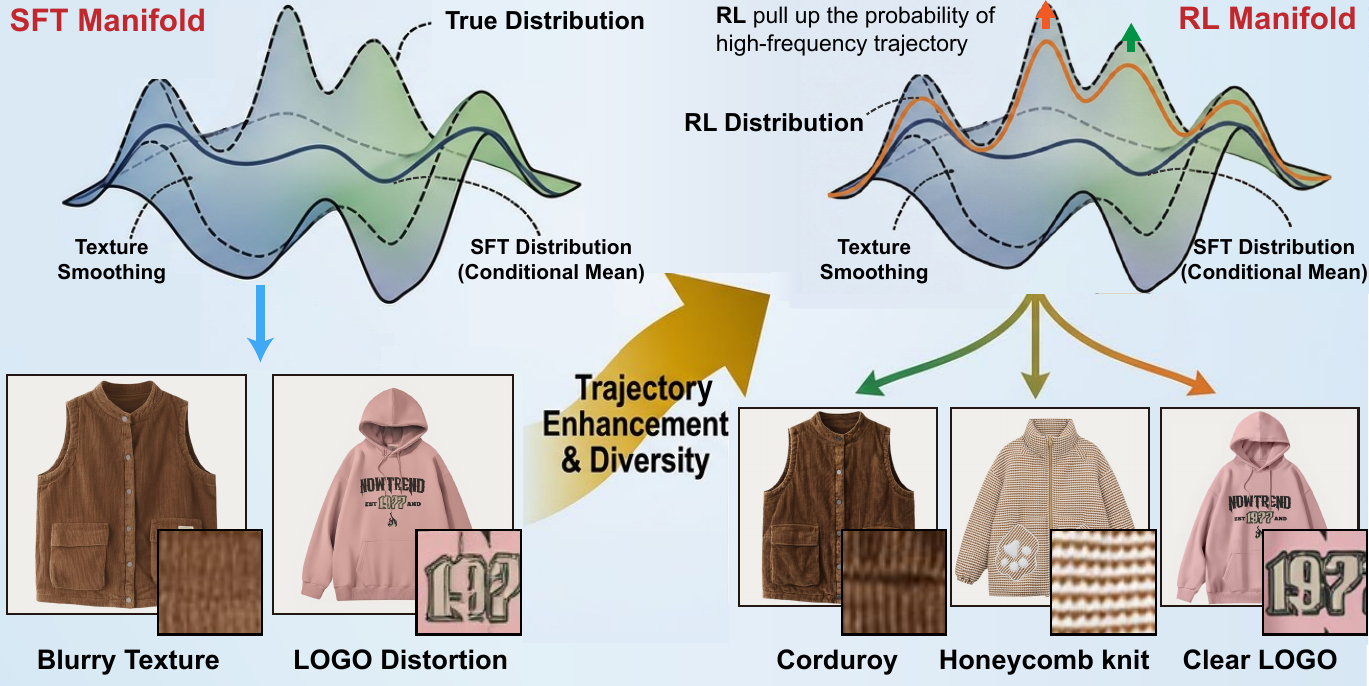}
  \caption{\textbf{High-Frequency Trajectory Enhancement via Distribution Reshaping.} \textbf{Left:} Current SFT-based models regress to the conditional mean, smoothing out textures. \textbf{Right:} RL elevates probability in true high-frequency modes, enabling various and high-fidelity sampling.}
\label{fig: moti_new}
\end{figure}

\IEEEPARstart{R}{ecent} years have witnessed breakthrough progress in text-to-image (T2I) diffusion models for general-purpose visual generation~\cite{banana, wu2025qwenimagetechnicalreport, liu2025step1x-edit}. However, industrial-grade fashion clothing asset synthesis~\cite{velioglu2024tryoffdiff, xarchakos2024tryoffanyone} (i.e., restoring high-specification flat-lay garment images from in-the-wild images) demands not only visual plausibility but also absolute fidelity to both the \textbf{macro-topology} and \textbf{micro-textures} of the reference images. In our preliminary work, RAGDiffusion~\cite{li2025ragdiffusion}, we effectively constrained the macroscopic geometry (e.g., sleeve length, silhouette) by introducing a Retrieval-Augmented Generation (RAG) paradigm~\cite{ram2023conrag2}, thereby eradicating large-scale structural hallucinations. Nevertheless, the final and most formidable barrier to commercial deployment remains: enabling the model to stably sample fully faithful garment assets with accurate fabric textures and patterns (e.g., high-frequency tweed textures, intricate brand logos).

\begin{figure}[t]
  \centering
  \includegraphics[width=1\linewidth]{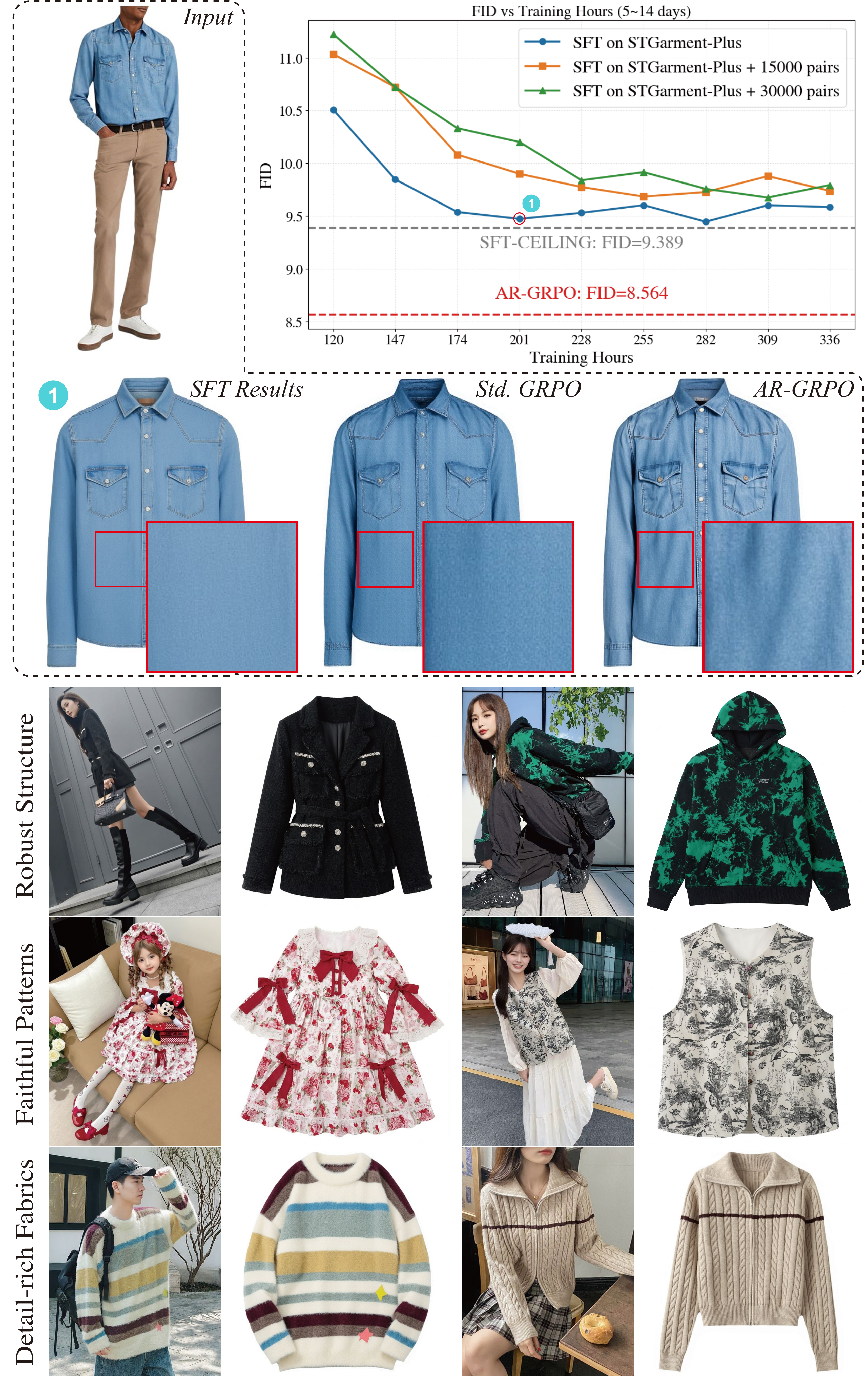}
    \caption{\textbf{Breaking the SFT saturation bottleneck with AR-GRPO.} During the SFT phase, the generative model reaches a performance plateau, while extending the training duration or scaling the data yields diminishing returns. The introduction of AR-GRPO triggers a sharp decline in FID, successfully breaking this bottleneck. The visual comparisons demonstrate SFT's limitations in capturing high-frequency physical details (e.g., complex fabric textures and intricate patterns) and the significant enhancements brought by RL optimization.}  \label{fig: motivation}
\end{figure}

\begin{figure*}[t]
  \centering
  \begin{minipage}{0.75\textwidth}
    \centering
    \includegraphics[width=\linewidth]{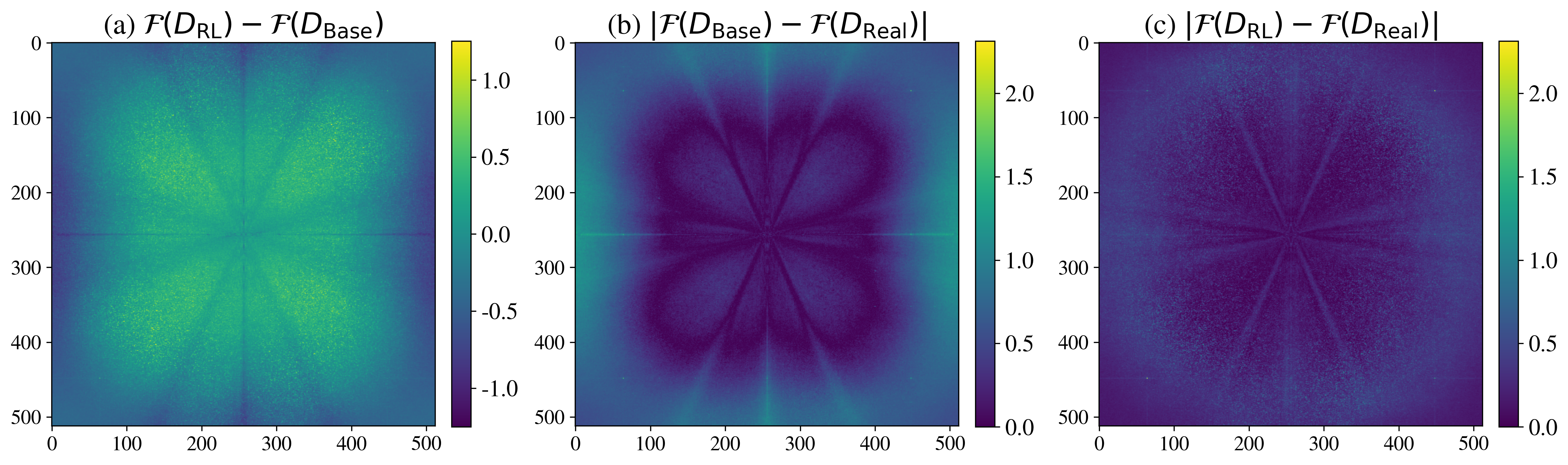}
    \centerline{(a) Frequency-domain comparison between generated and real images.}
  \end{minipage}
  \hfill
  \begin{minipage}{0.24\textwidth}
    \centering
    \includegraphics[width=\linewidth]{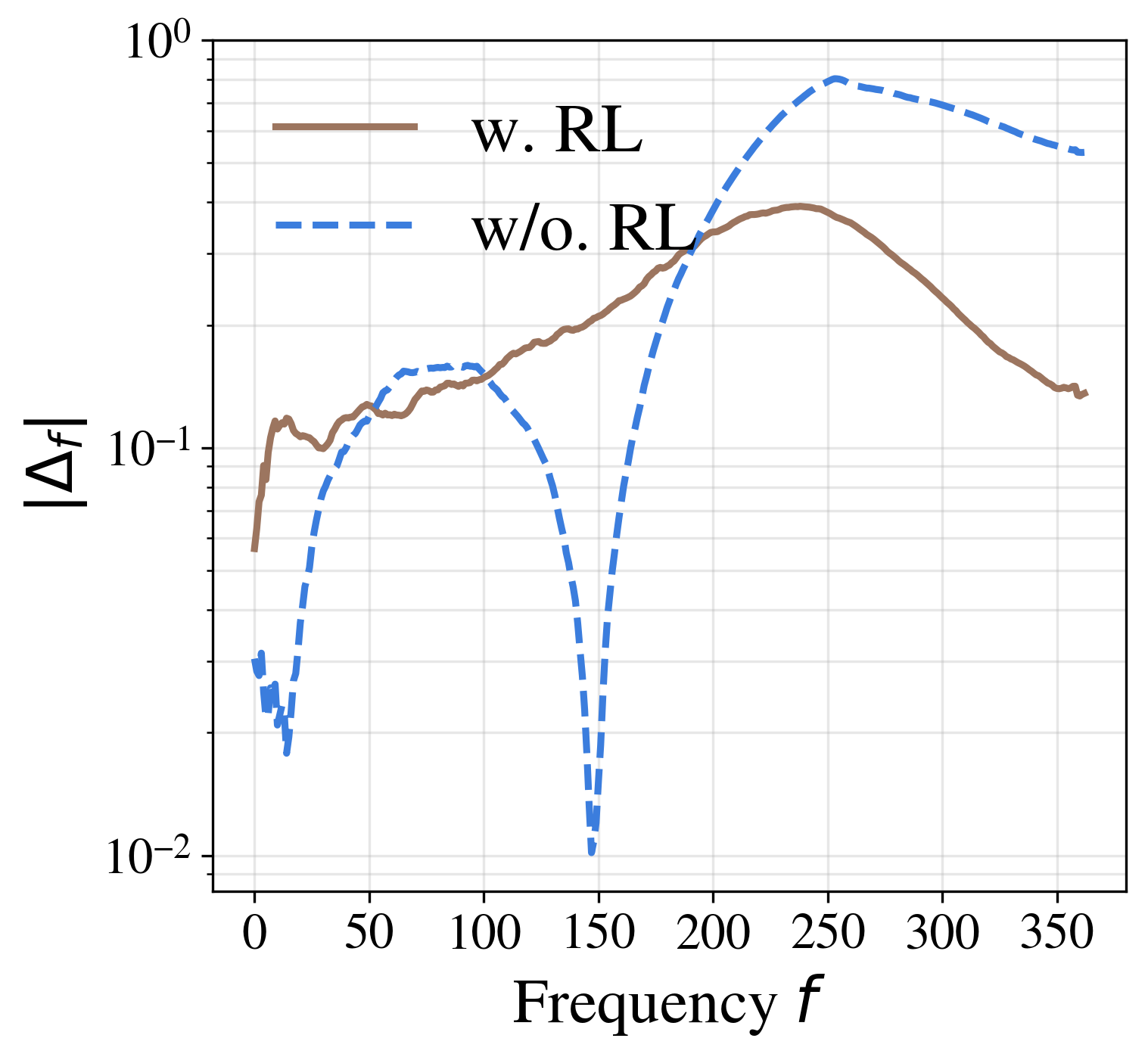}
    \centerline{(b) Radially spectral error curves.}
  \end{minipage}
  \caption{(a) shows Frequency-domain comparison between the base (SFT) model generation $D_{\mathrm{Base}}$, AR-GRPO results $D_{\mathrm{RL}}$, and real images $D_{\mathrm{Real}}$. Compared with the Base model, RL yields lower spectral discrepancy and weaker directional bias. (b) shows radially averaged spectral error curves with respect to the real-image spectrum. The RL model shows lower error over a broad middle-to-high frequency range.}
  \label{fig: freq}
\end{figure*}

To investigate this micro-level fidelity, we conducted empirical observations by generating dozens of candidates for the same condition using state-of-the-art supervised fine-tuning (SFT) flow models~\cite{flux}. Alarmingly, the results consistently collapsed into smooth fabrics with stochastic, blurry logos in our task. This empirical evidence demonstrates that faithful high-frequency expression is merely a ``rare'' event in the latent space;~\cite{berrada2024lpl, choi2025framer}, as SFT results in Fig.~\ref{fig: moti_new} and ~\ref{fig: motivation}. We define this fundamental probabilistic flaw as \textbf{High-Frequency Trajectory Collapse} (HFTC, frequency bias is illustrated in Fig.~\ref{fig: freq}). 
Conceptually, HFTC is the conditional generation manifestation of the well-known \textit{regression-to-the-mean} phenomenon in image restoration and super-resolution (SR)~\cite{ledig2017photorealisticsingleimagesuperresolution, Blau_2018}, where pixel-wise reconstruction objectives provably converge to the conditional expectation and thereby suppress high-frequency details. However, prior remedies---such as adversarial losses or perceptual constraints in SR---operate at the pixel level and cannot be trivially transferred to flow-matching-based Diffusion Transformers (DiTs), where the trajectory itself is the optimization target.  
As systematically categorized in Fig.~\ref{fig: moti_new} and ~\ref{fig: motivation}, this collapse manifests in multiple distinct failure modes: systematic texture smoothing where complex weaves degrade into flat plastic surfaces, logo dissolution, and pattern periodicity loss where regular stripes or checks become irregular.

We propose \textbf{AR-GRPO} (Adversarial-Regularized Group Relative Policy Optimization), a new \textit{Reinforcement Learning} (RL) post-training paradigm that performs \textbf{distribution reshaping with adversarial safeguard} at the trajectory level to fundamentally resolve High-Frequency Trajectory Collapse. Its effectiveness rests on the fact that SFT and RL operate on qualitatively different levels: SFT, via flow matching, regresses to the \textit{conditional mean} of the velocity field and is therefore confined to the smooth, low-frequency mode of the training distribution---which is precisely the regression-to-the-mean trap that produces HFTC. RL fundamentally departs from this objective: by replacing the deterministic Ordinary Differential Equation (ODE) inference of flow matching with Stochastic Differential Equation (SDE) sampling and reweighting trajectories with reward-driven policy gradients, it elevates low-probability but physically correct high-frequency patterns into the sampleable regime. In other words, rather than smoothing pixels post hoc, RL reshapes the trajectory-level probability mass at its source, which is the mechanism required to escape the conditional-mean attractor.

Therefore, for RL to succeed in resolving High-Frequency Trajectory Collapse, three prerequisites emerge as logical consequences of this single principle:
\begin{enumerate}
    \item \textbf{Inherent Capacity:} The generative foundation must possess the latent ability to sample high-fidelity textures---RL cannot amplify a probability of zero.
    \item \textbf{Perceptive Reward:} A precise reward function that genuinely comprehends fine-grained garment attributes (local craftsmanship, fabrics, material textures), rather than high-level semantic similarity.
    \item \textbf{Hacking Prevention:} A robust adversarial mechanism that detects and penalizes exploitation of reward blind spots, ensuring that RL-driven exploration enriches only authentic details.
\end{enumerate}

Obviously, current paradigms fail on all three fronts. \textbf{First}, existing garment generation models lack crucial material priors and high-complexity training data (Fig.~\ref{fig: main}). \textbf{Second}, existing generic rewards (e.g., PickScore~\cite{kirstain2023pick}, HPSv2~\cite{wu2023hpsv2}, DINO similarity~\cite{simeoni2025dinov3}) are inherently biased towards high-level semantic similarity---their human preference accuracy on garment evaluation falls below 50\%---lacking the capacity to comprehend fine-grained garment attributes. \textbf{Third}, when the reward signal contains blind spots, RL inevitably discovers and amplifies them and degrades the quality of generation, which has been studied as \textit{reward hacking} in language-model Reinforcement Learning~\cite{fu2026rewardshapingmitigatereward}. We term its specific incarnation in our setting \textbf{Artifact Hacking}: dense checkerboard noise, regular grid patterns, and high-frequency speckles that hijack the texture-insensitive blind spots of vision rewards to achieve spuriously high scores while degrading visual quality (Fig.~\ref{fig: reward_hack}). Without an explicit hacking-prevention mechanism, RL post-training degenerates into Artifact Hacking rather than authentic detail enrichment.

Centered on \textbf{AR-GRPO} as the core optimization strategy, we instantiate the distribution-reshaping paradigm into a unified framework, \textbf{RAGDiffusion++}, that is built upon the macro-constraints of our preliminary RAG~\cite{li2025ragdiffusion} and is equipped with three dedicated components, each satisfying one of the three prerequisites identified above. These components, together with the quantitative outcomes they deliver, constitute the core contributions of this work:

\begin{itemize}
    \item \textbf{Data \& Architecture Leap for Inherent Capacity} (Prerequisite~i): We construct \textbf{STGarment-Plus}, a 27,725-pair high-difficulty garment dataset focused on high-frequency materials, and upgrade the generative backbone from SDXL~\cite{podell2023sdxl} to a Dual-Image-Stream FLUX~\cite{flux} architecture for pure visual feature injection. This leap enriches the generative foundation's latent capability to sample high-fidelity textures---the very probability mass that AR-GRPO subsequently reshapes and amplifies.
    
    \item \textbf{Attribute-Aware Garment-RM for Perceptive Reward} (Prerequisite~ii): We pre-train \textbf{Garment-RM}, a fine-grained reward model, on 500K garment images via contrastive learning over 12 fine-grained attributes. It overcomes the semantic bias of generic vision foundation models, achieving 84.67\% human preference accuracy---far surpassing PickScore, HPSv2, and DINO similarity. Within RAGDiffusion++, Garment-RM thereby supplies AR-GRPO with an accurate, attribute-aware optimization direction, steering the trajectory-level distribution reshaping toward genuinely faithful textures rather than spurious semantic shortcuts.
    
    \item \textbf{Adversarial-Regularized GRPO for Hacking Prevention} (Prerequisite~iii): At the heart of RAGDiffusion++, AR-GRPO integrates a dynamic GAN discriminator~\cite{goodfellow2014GAN} into the RL sampling trajectory as a ``texture police'' that detects and penalizes low-level artifacts, while Garment-RM supplies the directional reward. This is the first work to formally expose the Artifact Hacking phenomenon in flow-based RL post-training and to neutralize it with adversarial regularization, ensuring that RL exploration enriches only authentic high-frequency details. 
\end{itemize}

Extensive experiments on all three benchmarks demonstrate that RAGDiffusion++ achieves consistent state-of-the-art performance, synthesizing garments that are simultaneously structurally and texturally faithful. Quantitatively, RAGDiffusion++ reduces FID by 14.3\% and KID by 34.2\% over our preliminary RAGDiffusion, surpasses the best open-source Qwen-Image-Edit baseline by 51.0\% in KID, and outperforms the large-scale commercial model Nano Banana by 10.9\% in FID. The full STGarment-Plus dataset and Garment-RM will be open-sourced upon acceptance.

\section{Related works}

\noindent \textbf{Controllable text-to-image diffusion models.} 
To attain conditional control in text-to-image diffusion models, ControlNet~\cite{zhang2023controlnet}, T2I-Adapter~\cite{mou2024t2iadapter}, and IP-Adapter~\cite{ye2023ipadapter} incorporate additional trainable modules to fuse conditions on feature maps. Additionally, recent investigations have utilized a variety of prompt engineering techniques~\cite{li2023gligen, yang2023reco, zhang2023controllablegpt4} and implemented cross-attention constraints~\cite{chen2024layoutcontrol, xie2023boxdiff, ju2023humansd, zhang2024boow} to facilitate more controllable generative processes. However, these methods primarily rely on the generic associative capabilities of pre-trained diffusion backbones. While effective for general stylistic or spatial control, they struggle to guarantee absolute pixel-level fidelity when synthesizing high-frequency physical materials (e.g., complex fabric weaves or intricate craftsmanship). Relying on such generic associative generation inevitably leads to texture over-smoothing and the catastrophic loss of high-frequency physical fidelity.

\noindent \textbf{Garment restoration.}  
Standard clothing generation involves restoring flat-lay garment images from real-world contexts. TileGAN~\cite{zeng2020tilegan} pioneered a two-stage GAN~\cite{goodfellow2014GAN} framework, suffering from backbone capability limitations. Recent works including TryOffDiff~\cite{velioglu2024tryoffdiff}, TryOffAnyone~\cite{xarchakos2024tryoffanyone}, and IGR~\cite{shen2024igr} have adopted pretrained SD as base backbones, yielding obvious improvements through enhanced generative priors. While our preliminary work RAGDiffusion~\cite{li2025ragdiffusion} addressed the structural inaccuracies ignored by these methods, existing frameworks still face the formidable \textbf{High-Frequency Trajectory Collapse}. Devoid of high-frequency physical priors, they fail to render authentic micro-materials (e.g., tweed, intricate logos), typically producing an over-smoothed ``AI-generated plastic feel''.

\noindent \textbf{Reinforcement learning in image generation.} 
Reinforcement learning (RL) has recently emerged as a powerful paradigm to align generative models with human preferences, moving beyond standard maximum likelihood estimation. Foundational works such as DDPO~\cite{black2023ddpo} and DPOK~\cite{fan2023dpok} first formulated the iterative denoising process of diffusion models as a Markov Decision Process (MDP), enabling the direct optimization of non-differentiable rewards like ImageReward~\cite{xu2023imagereward} or Aesthetic scores. With the evolution towards continuous-time flow models, recent studies have adapted more advanced RL algorithms. For instance, Flow-GRPO~\cite{liu2025flowgrpo} extends group relative policy optimization to normalizing flows, while DANCE~\cite{xue2025dancegrpo} and Mix-GRPO~\cite{li2025mixgrpo} focus on blending multiple reward signals for comprehensive prompt-following. Additionally, SRPO~\cite{shen2025srpo} introduces step-wise preference optimization to enhance visual appeal. While these methods achieve remarkable success in generic text-to-image enhancements, they predominantly rely on generic reward models (e.g., PickScore~\cite{kirstain2023pick}, HPSv2~\cite{wu2023hpsv2}) that are inherently biased towards high-level semantic alignment. When applied to industrial-grade physical rendering, these semantic-biased rewards lack the granularity to supervise high-frequency material details (e.g., complex fabric weaves). Consequently, directly applying these generic RL frameworks to garment synthesis fails to resolve the \textit{High-Frequency Trajectory Collapse} and possibly triggers \textbf{Artifact Hacking}, where models generate deceptive checkerboard noise to exploit reward vulnerabilities. In contrast, our framework introduces a specialized Garment-RM and an adversarial GRPO strategy to ensure RL optimization strictly adheres to physical fidelity.  
\begin{figure*}[t]
  \centering
  \includegraphics[width=0.95\linewidth]{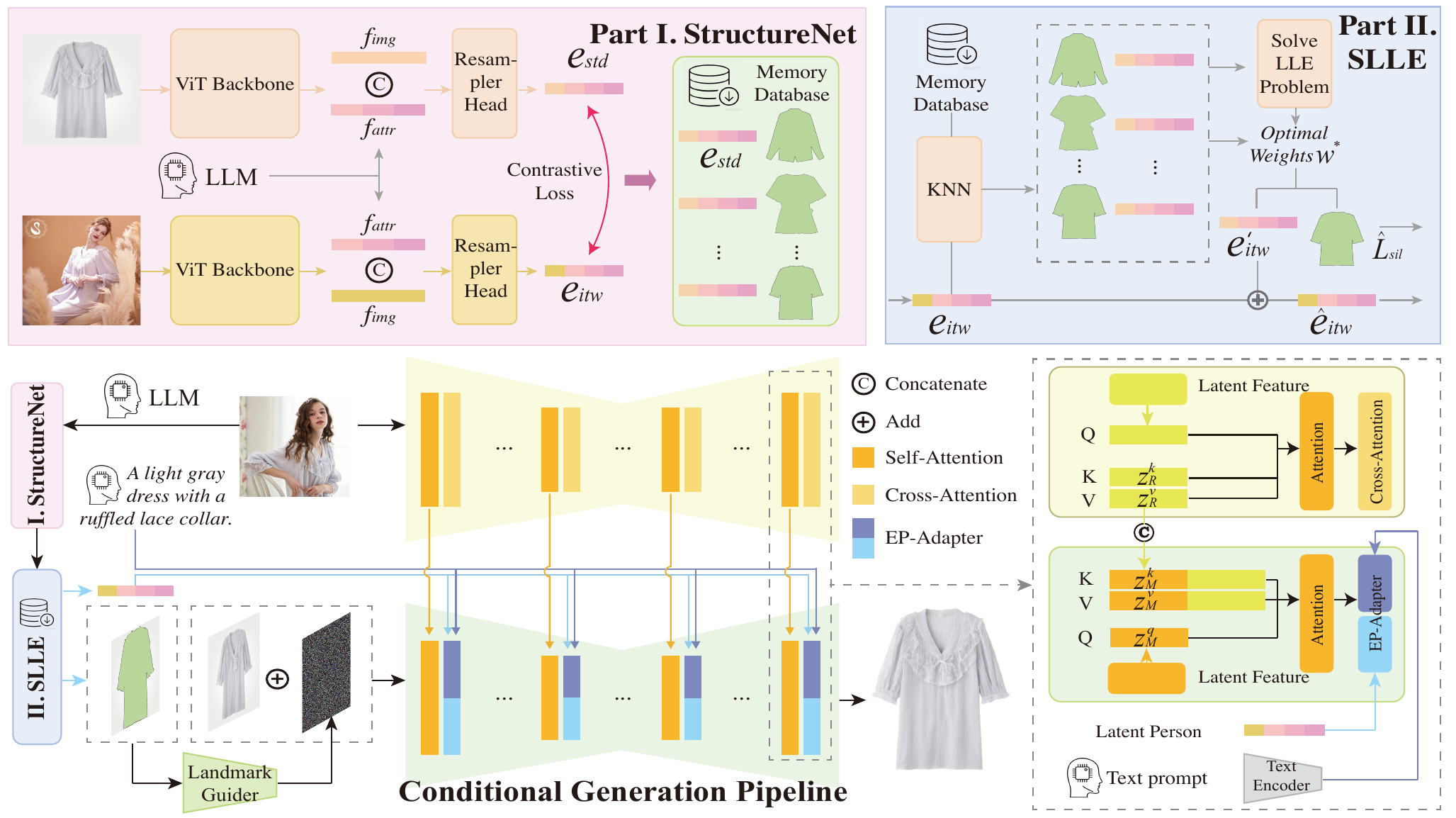}
  \caption{Overall framework of our previous work RAGDiffusion. StructureNet provides latent structural embeddings, while SLLE facilitates the embedding fusion along with landmark retrieval. The generative model synthesizes multiple conditions to achieve omni-level high-fidelity generation.  }
  \label{fig: old_framework}
\end{figure*}

\section{Preliminary Work Revised: RAGDiffusion}
\label{sec: preliminary}

Before delving into the micro-alignment reinforcement learning paradigm proposed in this paper, we first briefly revisit our preliminary work, RAGDiffusion~\cite{li2025ragdiffusion}, which serves as the macro-topological constraint of our system. 

Unlike conventional virtual try-on paradigms~\cite{xu2024ootdiffusion, li2024anyfit}, industrial-grade garment synthesis is a highly demanding and sampling-constrained process. It necessitates strict standard structures and frontal positioning, which intrinsically limit the stylistic flexibility and confines the generative sampling to a narrow manifold of high-specification garment distributions. Furthermore, the profound complexity of in-the-wild conditions---such as severe occlusions, multi-layer outfits, and extreme lateral viewpoints---notably exacerbates structural uncertainty and pattern ambiguity. Consequently, existing methods fail to achieve practical usability thresholds. They typically suffer from pathological \textit{structure hallucinations}~\cite{li2025ragdiffusion}, manifesting as severely inaccurate structure, fitness, and length estimations under challenging scenes.  

To circumvent these limitations, RAGDiffusion introduces a Retrieval-Augmented Generation (RAG) paradigm~\cite{ram2023conrag2} that assimilates explicit external priors to anchor macroscopic geometry and eradicate structural hallucinations.
An overview of the RAGDiffusion is presented in Fig.~\ref{fig: old_framework}. The backbone employs SDXL~\cite{rombach2022ldm}. A dual tower StructureNet extracts latent structure embeddings by contrastive learning as detailed in Sec. \ref{sec: emb}, and SLLE retrieves and fuses structure embeddings and landmarks as in Sec. \ref{sec: SLLE}. The conditional generative pipeline is described in Sec. ~\ref{sec: omni}.

\subsection{Dual-tower embeddings extraction}
\label{sec: emb}

To eliminate background interference and force the model to focus on the pure physical structure of garments, we designed a dual-tower StructureNet, denoted as $g_{\theta}$, to extract latent structural embeddings. We adopt a contrastive learning paradigm~\cite{radford2021clip} to capture structural consistency. Specifically, during training, we construct batches of $N$ image pairs consisting of in-the-wild clothing $x_{itw}$ and standard flat-lay clothing $x_{std}$. Positive pairs are formulated using garments that share similar edge contours (Canny responses) but possess drastically different textures. 

To comprehensively capture the structural semantics and eliminate the semantic deficiencies and ambiguities present in pure vision, we leverage a Large Language Model (LLM)~\cite{achiam2023gpt, bai2023qwen} to extract 10 types of discrete structural attributes (e.g., Category, FitType, CollarTechnique). We then assign a learnable embedding $f_{attr}$ to each discrete attribute. Simultaneously, the visual structure features $f_{img}$ are extracted by a twin ViT~\cite{dosovitskiy2020vit} encoder. These visual features $f_{img}$ are concatenated with the attribute embeddings $f_{attr}$ and processed through a non-linear Resampler~\cite{ye2023ipadapter} head to formulate the final latent structure embeddings $e \in \{e_{itw}, e_{std}\}$.

Optimized via the InfoNCE loss~\cite{chen2020simclr}, StructureNet learns a robust multi-modal embedding space by maximizing the cosine similarity of the $N$ matched positive pairs (denoted with superscript $+$) while pushing apart the $N^2-N$ structurally dissimilar incorrect pairs (denoted with superscript $-$). The objective function is formulated as:
\begin{equation}
\small
\mathcal{L} = \frac{-1}{N}\sum_{i=1}^{N}\text{log}\frac{\text{exp}(e_{itw, i}\odot e_{std}^+/\tau)} {\text{exp}(e_{itw, i}\odot e_{std}^+/\tau)+ {\textstyle \sum_{e_{std}^-}} \text{exp}(e_{itw, i}\odot e_{std}^-/\tau)},
\end{equation}
where $\odot$ denotes the cosine similarity between two vectors, $N$ is the batch size, and $\tau$ is a temperature scalar. Through this rigorous alignment, given a complex in-the-wild garment image $x_{itw}$ during inference, this module robustly extracts a clean structural embedding $e_{itw}$ with environmental noise and texture variations thoroughly filtered out.


\subsection{Retrieval-based structure aggregation}
\label{sec: SLLE}

Due to the limited spatial perception inherent in the SD~\cite{borji2023qualitative, liu2023intriguing}, it often struggles to accurately represent the length and contours of clothing, particularly in hard cases (occlusions, multi-layer, lateral viewpoints, etc). Furthermore, relying solely on StructureNet may yield out-of-distribution (OOD) errors for extreme in-the-wild samples. To circumvent this, we establish an external feature memory database $\mathcal{D}$. This database stores a massive collection of high-quality standard flat-lay structural embeddings $e_{std}$ and their precisely corresponding silhouette landmarks $L_{sil}$, formulated as retrieval pairs $(e_{std}, L_{sil})$.

To seamlessly map the potentially noisy features $e_{itw}$ extracted from in-the-wild scenes onto the reliable feature manifold of standard garments, we propose the \textbf{Structure Locally Linear Embedding (SLLE)} algorithm~\cite{roweis2000lle, blanz2003face}. This retrieval-augmented process operates in two steps to mathematically anchor the macroscopic skeleton. First, given an extracted in-the-wild cloth query embedding $e_{itw}$, we utilize cosine similarity to retrieve the $K$-nearest standard reference embeddings $\{e_{std}^1, \dots, e_{std}^K\}$ and their paired silhouette landmarks $\{L_{sil}^1, \dots, L_{sil}^K\}$ from the database $\mathcal{D}$~\cite{guo2003knn}.

Subsequently, assuming the local manifold is approximately linear, SLLE calculates the optimal barycentric weights for a locally linear combination by minimizing the reconstruction error. This is formulated as a least-squares optimization problem:
\begin{equation}
\min_{w} \left\| e_{itw} - \sum_{i=1}^{K} w_i \cdot e_{std}^i \right\|_2, \quad \text{s.t.} \sum_{i=1}^{K} w_i = 1.
\label{eq:slle_opt}
\end{equation}
Solving Eq.~\ref{eq:slle_opt} yields the optimal weights $\{w_1^*, \dots, w_K^*\}$. To mitigate information loss while forcefully pulling the OOD $e_{itw}$ back to the in-domain target space, the final rectified structure representation $\hat{e}_{itw}$ is obtained through a linear fusion of the original embedding and the reconstructed one:
\begin{equation}
\hat{e}_{itw} = \alpha \cdot \sum_{i=1}^{K} w_i^* \cdot e_{std}^i + (1-\alpha) \cdot e_{itw},
\label{eq:slle_fuse}
\end{equation}
where $\alpha \in [0, 1]$ controls the trade-off and is empirically set to $0.5$. The final landmark $\hat{L_{sil}}$ is also fused with optimal weights. Through this precise local reconstruction, the SLLE algorithm effectively eliminates structural outliers and mathematically guarantees macroscopic geometric fidelity before the generation phase.


\subsection{Conditional generative pipeline}
\label{sec: omni}

Upon acquiring the rectified structural embedding $\hat{e}_{itw}$ and silhouette landmark $\hat{L}_{sil}$ from the retrieval module, the preliminary RAGDiffusion integrates these priors into the generation pipeline to strictly constrain the macro-topology. Specifically, an Embedding Prompt Adapter (EP-Adapter)~\cite{ye2023ipadapter} is employed to inject $\hat{e}_{itw}$ via parallel cross-attention layers, acting as a soft semantic constraint. Concurrently, a spatial Landmark Guider~\cite{hu2023animate} processes the silhouette mask $\hat{L}_{sil}$ to align with the latent space as an explicit hard constraint. 

To further preserve the coarse patterns of the garment, an isomorphic ReferenceNet~\cite{hu2023animate} is utilized. It extracts intermediate key and value features $\left \{ z_{R}^k, z_{R}^v \right \}$ from the in-the-wild garment image. These are concatenated with the MainNet features $\left \{ z_{M}^k, z_{M}^v \right \}$ along the sequence dimension to form $\left \{ z_{C}^{k}, z_{C}^{v} \right \}$, which then guides the self-attention mechanism:
\begin{equation}
\text{Attention}\left( z_{M}^q, z_{C}^k, z_{C}^v \right) = \text{softmax}\left(\frac{z_{M}^q z_{C}^{kT}}{\sqrt{d}}\right)z_{C}^v,
\end{equation}
where $z_{M}^q$ represents the query features from the MainNet. This attention sharing plays a crucial role in preserving the basic patterns of complicated garments.

\noindent \textbf{Limitations of the Preliminary Pipeline.} 
While the aforementioned SDXL-based pipeline effectively eradicated macroscopic structural hallucinations, it encountered a bottleneck in micro-detail faithfulness (e.g., intricate brand logos and high-frequency fabrics). This degradation was partly exacerbated by the severe information loss of the SDXL VAE. In RAGDiffusion, we attempted to heuristically mitigate this by proposing a Parameter Gradual Encoding Adaption (PGEA) module to forcefully align the SDXL UNet with a higher-capacity VAE. 

However, this patching strategy remained bounded by the intrinsic frequency expression bottleneck of the diffusion architecture (in Fig.~\ref{fig: motivation} and Fig.~\ref{fig: freq}). It merely alleviated decoding loss but failed to fundamentally avert the \textit{High-Frequency Trajectory Collapse} during the denoising sampling. Recognizing that heuristic patches cannot cure foundational generative flaws, in this work, we introduce a training paradigm leap to an adversarial RL framework, complemented by a native pure-visual Dual Image-Stream DiT, to establish a definitive solution for micro-texture fidelity. These limitations directly motivate the three prerequisites for micro-fidelity RL post-training---inherent capacity, perceptive reward, and hacking prevention---that we systematically address in the following section.

\section{Method}

\begin{figure*}[t]
  \centering
\includegraphics[width=1.0\textwidth]{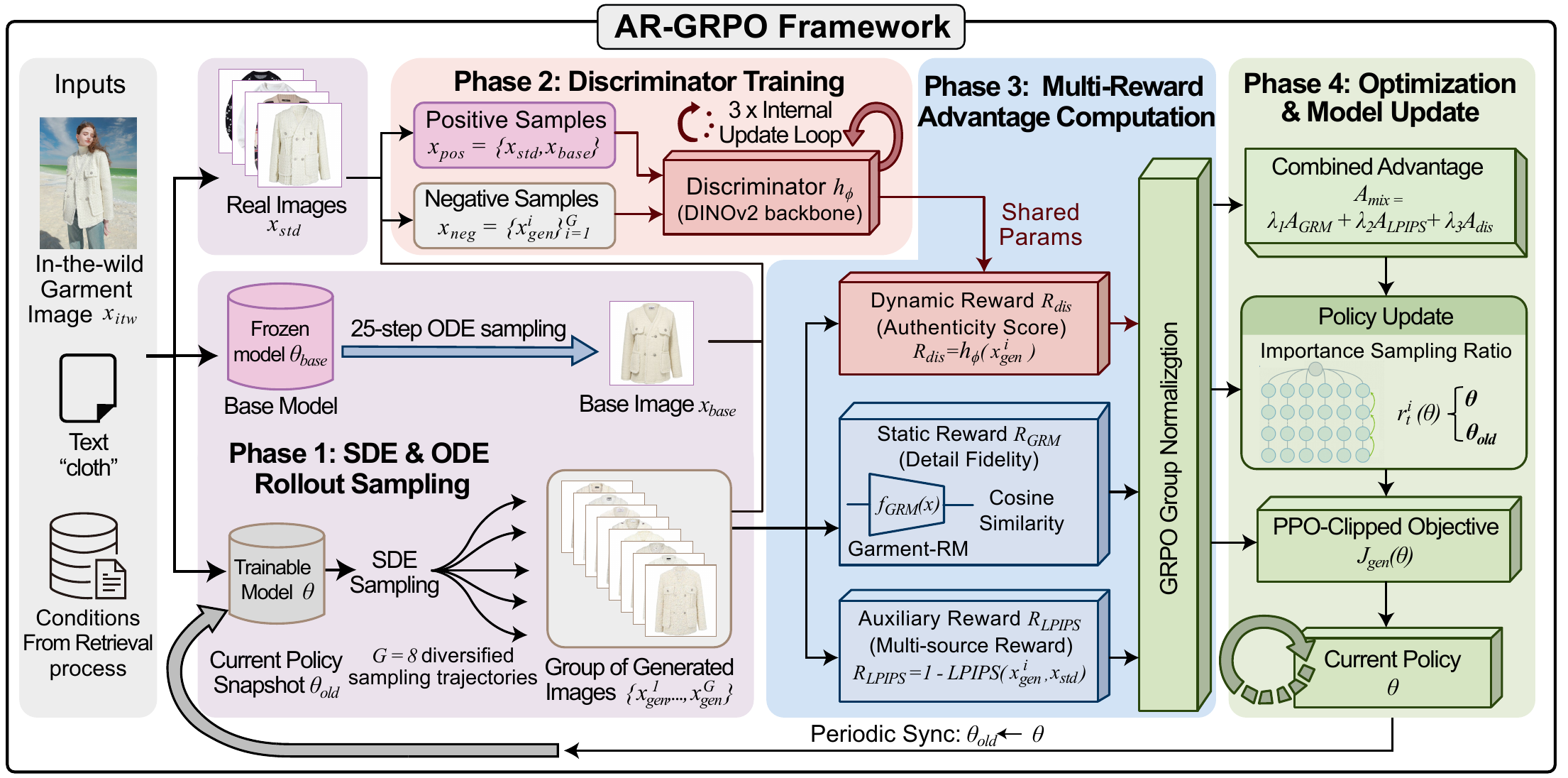}
    \caption{AR-GRPO training framework consists of four phases: Phase 1 generates diverse trajectories via SDE sampling; Phase 2 trains the discriminator $h_\phi$ adversarially to establish an authenticity reward; Phase 3 computes combined advantages from Garment-RM, LPIPS, and discriminator; Phase 4 updates the generative policy. This adversarial regularization ensures the model achieves high fidelity while avoiding artifacts.} 
    \label{fig: framework_grpo}
\end{figure*}

\begin{figure*}[t]
  \centering
\includegraphics[width=1\textwidth]{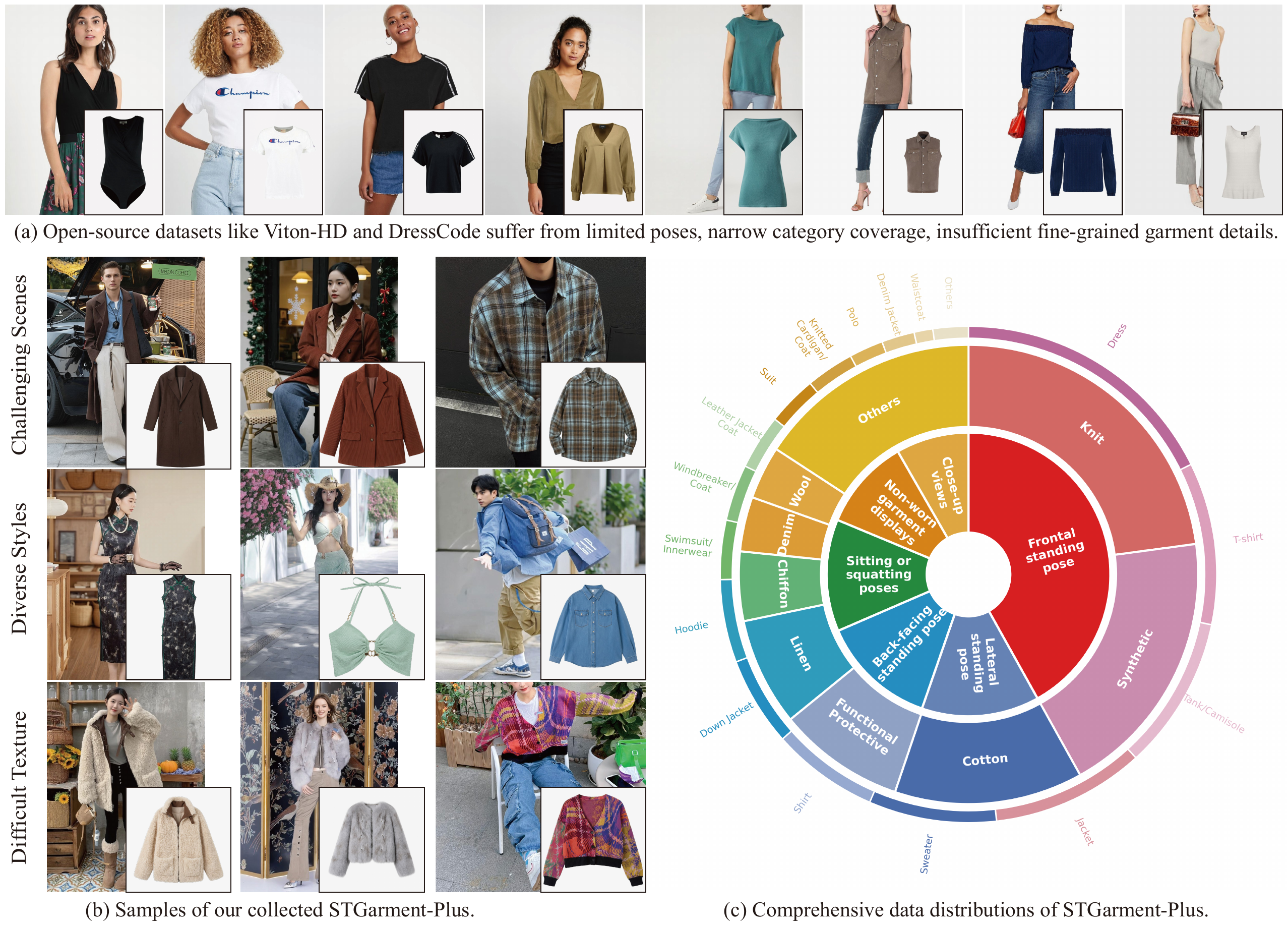}
  \caption{Overview of STGarment-Plus. (a) and (b) give a visual comparison of STGarment-Plus and open-source datasets. (c) STGarment-Plus covers challenging real-world scenes, diverse garment styles, and difficult high-frequency textures.}
  \label{fig: dataset_comare}
\end{figure*}

\begin{figure}[t]
  \centering
\includegraphics[width=1.0\linewidth]{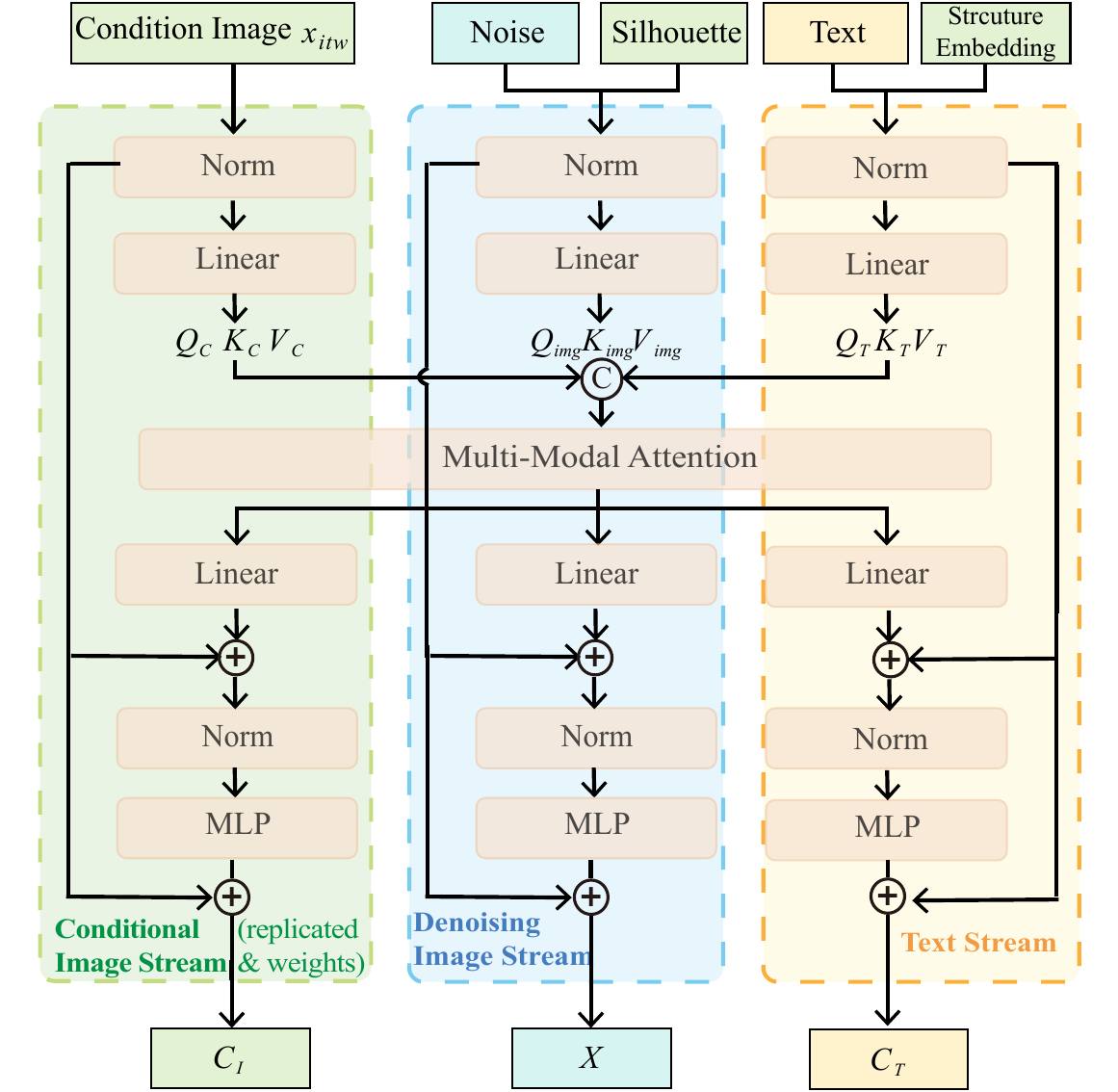}
  \caption{Dual-Image-Stream DiT framework. Our model introduces a dedicated conditioning image stream by replicating part of the denoising image stream and its pretrained weights.} 
  \label{fig: framework_dit}
\end{figure}

Given an in-the-wild clothing condition image $x_{itw} \in \mathbb{R}^{H\times W\times 3}$, RAGDiffusion++ aims to synthesize an authentic standard flat-lay garment image $x_{std}$ that strictly preserves both macroscopic topology and microscopic physical materials. Before detailing each component, we articulate the core insight that unifies our framework.

\noindent \textbf{Core Insight: RL as Distribution Reshaping.}
SFT via flow matching trains the model to predict the \textit{conditional mean} of the velocity field, which by definition converges to the mode of the training distribution. In garment generation, this mode is dominated by smooth, low-frequency textures---statistically more prevalent and robust to denoising noise---while high-frequency patterns (tweed weaves, intricate logos, knitted ribs) occupy a thin distributional tail. Since SFT employs deterministic ODE inference, the same noise always produces the same smoothed output; the model is structurally incapable of exploring beyond this modal collapse---the probabilistic root cause of \textit{High-Frequency Trajectory Collapse} (Fig.~\ref{fig: freq}). RL breaks this trap through a qualitatively different mechanism: replacing ODE with stochastic SDE sampling enables trajectory exploration beyond the SFT mode, and reward-weighted policy gradients then systematically up-weight trajectories that produce high-frequency-faithful outputs, effectively elevating low-probability but physically correct patterns into the sampleable regime. However, this distribution reshaping is only as good as the reward signal---an imprecise or semantically biased reward leads the model to exploit reward blind spots, triggering \textit{Artifact Hacking}. Three prerequisites therefore emerge as logical consequences:

\begin{itemize}
    \item \textbf{Inherent Capacity} (Sec.~\ref{sec: dataset and architecture}): RL cannot amplify a probability of absolute zero. STGarment-Plus dataset and a Dual-Image-Stream FLUX architecture establish non-zero high-frequency texture priors.
    \item \textbf{Perceptive Reward} (Sec.~\ref{sec: rm}): Garment-RM provides accurate, attribute-aware reward signals that genuinely perceive material-level quality.
    \item \textbf{Hacking Prevention} (Sec.~\ref{sec: ar_grpo}): An adversarial discriminator detects and penalizes artifacts, ensuring RL exploration enriches only authentic details.
\end{itemize}

\noindent The overall AR-GRPO pipeline is illustrated in Fig.~\ref{fig: framework_grpo}.

\subsection{SFT: Data Improvement and Architecture Update}
\label{sec: dataset and architecture}

As established in the insight above, RL can only amplify sampling probabilities that are already non-zero. The SFT phase must therefore ensure the model possesses latent capacity for high-frequency texture generation, serving as the \textit{inherent capacity} prerequisite. We address this through two complementary strategies: curating a challenging high-frequency dataset and upgrading the architecture.

\subsubsection{STGarment-Plus Dataset Construction}
\label{sec: dataset}

Standard garment restoration requires paired data consisting of a clean flat-lay target and its corresponding in-the-wild reference. Existing virtual try-on benchmarks, such as VITON-HD and DressCode, have been widely adopted for this purpose, but they mostly only contain forward-facing persons under clean backgrounds with limited occlusion, narrow category coverage, simple materials, and insufficient fine-grained garment details. As a result, models trained on such sanitized data often suffer from a severe domain gap in real-world restoration. To mitigate this issue, our preliminary RAGDiffusion collected \textit{STGarment}, a web-scale dataset of about 65,131 pairs, which provides broader pre-training coverage but is still heavily affected by inconsistent quality, including low resolution, watermarks, and text overlays. While useful for early-stage pre-training, such noise limits the effectiveness of subsequent SFT and RL optimization. Therefore, we further curate \textbf{STGarment-Plus}, a high-resolution and category-comprehensive dataset specifically designed for garment restoration under complex textures, high-frequency materials, and challenging in-the-wild conditions, as shown in Fig.~\ref{fig: dataset_comare}

\noindent \textbf{Data Processing and Synthesis Pipeline.}
STGarment-Plus is built from two sources: a rigorously cleaned subset of the original STGarment and a curated synthetic subset generated by Nano Banana Pro~\cite{banana}. Starting from mixed raw image collections, we first use Qwen-VL~\cite{bai2023qwen} to separate standard flat-lay garments from in-the-wild references and remove near-duplicate samples. We then segment garment regions, extract garment features with our pre-trained Garment-RM (Sec.~\ref{sec: rm}), and filter mismatched pairs by feature similarity, followed by manual cleaning to remove low-quality samples, watermarks, and text overlays. To address long-tail gaps in rare categories and difficult fabrics, we further retrieve high-quality in-the-wild images and synthesize paired flat-lay garments with Nano Banana, followed by strict human verification across five strict dimensions: structural consistency, visual realism, background cleanliness, robustness against extreme occlusion, and absolute brand logo fidelity. This pipeline yields a high-resolution and reliable paired dataset for garment restoration.

\noindent \textbf{Comprehensive Data Distributions.} 
STGarment-Plus comprises 25,756 training pairs and 1,969 test pairs (27,725 in total) and exhibits substantial diversity in scenes, category, and material distributions. As shown in Fig.~\ref{fig: dataset_comare} (c), the dataset abundantly encapsulates challenging, real-world conditions, including sitting, squatting, lateral pose, and half-body close-ups with severe truncations. In terms of category coverage, over 50\% of the garments belong to complex structures (e.g., Down Jacket, Innerwear, Coat) that are entirely absent from open-source academic benchmarks. Most crucially, STGarment-Plus is exceptionally rich in diverse, high-frequency fabrics (such as delicate tweed, dense knits, and textured leather). These difficult materials effectively expose the high-frequency trajectory collapse in existing generative models, thereby providing an optimal, high-difficulty testbed and an indispensable training foundation for our micro-alignment paradigm. To facilitate future research in garment generation and related fine-grained synthesis tasks, we will publicly release the full STGarment-Plus dataset upon acceptance.

\subsubsection{DiT Framework and Training Object}
\label{sec: dual-stream}
FLUX~\cite{flux} is a text-to-image model built on stacked Multi-Modal Diffusion Transformers (DiT) blocks with separate text and image streams, demonstrating a significantly higher performance ceiling than the SDXL backbone. While Prior tuning methods~\cite{tan2024ominicontrol,wang2025unicombine} adapt such DiT models by directly concatenating condition image tokens with noisy image and text tokens, we customize the FLUX backbone into a \textbf{Dual-Image-Stream DiT} framework by replicating a portion of the denoising stream's architecture and pre-trained weights to initialize a dedicated \textit{conditioning image stream}, as illustrated in Fig.~\ref{fig: framework_dit}.

Following the above architectural modification, we train the proposed Dual-Image-Stream DiT using the standard Flow Matching objective~\cite{lipman2022flow, flux}. Given the target garment latent $z_0=\mathcal{E}(x_{std})$ and Gaussian noise $z_1\sim\mathcal{N}(0,I)$, we form the interpolated latent $z_t=(1-t)z_0+t z_1$ and supervise the model to predict the velocity field:
\begin{equation}
\mathcal{L}_{SFT} = \mathbb{E}_{t \sim \mathcal{U}(0,1), z_1 \sim \mathcal{N}(0,I), x_{std}} \left[ \left\| v_\theta(z_t, t, c) - (z_1 - z_0) \right\|_2^2 \right],
\label{eq:fm_loss}
\end{equation}
where $v_\theta$ is the predicted velocity field and $c$ denotes the visual conditioning signals derived from the reference garment and retrieval priors.


\subsection{Garment-RM: Attribute-Aware Garment Reward}
\label{sec: rm}

The distribution-reshaping principle (Sec.~\ref{sec:intro}) demands a reward function that genuinely perceives fine-grained material quality---the \textit{perceptive reward} prerequisite. However, existing generic vision foundation models (e.g., CLIP~\cite{radford2021clip} and DINO~\cite{simeoni2025dinov3}) are inherently biased towards high-level semantics: they assess ``does this look like a shirt?'' rather than ``is this tweed texture physically correct?'' As quantified in Tab.~\ref{tab: human_preference_acc}, widely used aesthetic reward models (HPSv2~\cite{wu2023hpsv2}, PickScore~\cite{kirstain2023pick}) achieve human preference accuracy \textit{below 50\%} on garment evaluation---worse than random selection.

\noindent \textbf{Design Principle.}
Our key observation is that garment quality is fundamentally an \textit{attribute-level} concept: a garment with the correct collar but wrong fabric is qualitatively different from one with both correct. Therefore, the reward model must learn a feature space where individual material attributes are explicitly disentangled, enabling it to provide precise, fine-grained optimization directions rather than holistic semantic similarity. To this end, we develop \textbf{Garment-RM}, a specialized reward model trained through fine-grained attribute classification and instance-level contrastive learning.

\noindent \textbf{Consensus-Driven Data Engine.}
To build Garment-RM without costly manual annotation, we construct a dual-VLM consensus data engine. We first define 12 fine-grained garment attributes, covering material, surface texture, local decorations, and other detailed properties. We then use GPT-4o and Qwen3-VL-8B to independently annotate a large collection of in-the-wild and flat-lay garment images, and retain only samples with consistent predictions from both models, yielding about 500,000 reliable training images. In addition, we mine the same garment under different conditions to form triplets of anchor ($x_a$), positive ($x_p$), and negative ($x_n$) samples for instance-level contrastive learning.

\noindent \textbf{Multi-Task Architecture and Objective.} 
Garment-RM employs a lightweight, EfficientNet-style~\cite{tan2019efficientnet} backbone equipped with Squeeze-and-Excitation (SE) modules~\cite{hu2018squeeze} to extract hierarchical visual features. The network bifurcates into a multi-task head: 12 parallel classification heads for fine-grained attributes and one global embedding head yielding a 256-dimensional continuous feature vector. 

To maximize the discriminative capacity of the model against subtle material differences, the training objective is formulated as a joint multi-task loss:
\begin{equation}
\mathcal{L}_{RM} = \lambda_{cls} \mathcal{L}_{cls} + \lambda_{tri} \mathcal{L}_{tri}.
\label{eq: rm_total_loss}
\end{equation}

For the attribute classification branch, we utilize the Additive Margin Softmax (AM-Softmax) loss~\cite{wang2018additive} $\mathcal{L}_{cls}$ to explicitly pull features of the same attribute closer while pushing disparate attributes apart. Given the L2-normalized feature embedding $x_i$ and the normalized weight vector $W_j$ of the last fully connected layer, the AM-Softmax loss for a batch of $N$ samples is formulated as:
\begin{equation}
\mathcal{L}_{cls} = -\frac{1}{N} \sum_{i=1}^{N} \log \frac{e^{s(\cos\theta_{y_i} - m)}}{e^{s(\cos\theta_{y_i} - m)} + \sum_{j=1, j \neq y_i}^{C} e^{s \cos\theta_j}},
\label{eq: am_softmax}
\end{equation}
where $C$ is the number of classes for a specific attribute, and $\theta_j$ denotes the angle between $W_j$ and $x_i$. The hyperparameter $s$ scales the cosine values, while the additive margin $m$ imposes a stricter angular penalty on the ground-truth class $y_i$. This mathematically forces the network to learn highly compact intra-class and maximally separated inter-class feature distributions.

Simultaneously, the global embedding branch is also supervised by a margin-based Triplet Loss $\mathcal{L}_{tri}$. Given the mined triplet $(x_a, x_p, x_n)$ and the extracted L2-normalized embeddings $(f_{GRM}(x_a), f_{GRM}(x_p), f_{GRM}(x_n))$, the triplet loss enforces instance-level fidelity in the angular space:
\begin{equation}
\begin{aligned}
\small
\mathcal{L}_{tri} = \max \Bigl( 0,\; & \bigl(f_{GRM}(x_a) \odot f_{GRM}(x_n)\bigr) \\
&- \bigl(f_{GRM}(x_a) \odot f_{GRM}(x_p)\bigr) + \alpha \Bigr),
\end{aligned}
\label{eq:triplet_loss}
\end{equation}
where $\odot$ denotes the cosine similarity, $f_{GRM}(\cdot)$ denotes the embedding extracted by the Garment-RM, and $\alpha$ is the predefined triplet margin.

\noindent \textbf{Reward Formulation.} 
The pre-trained Garment-RM acts as a garment-aware evaluator of generative quality. For a generated flat-lay garment $x^i_{gen}$ and its flat-lay ground-truth reference $x_{std}$, the continuous reward signal is explicitly formulated as their feature cosine similarity:
\begin{equation}
\label{eq: reward_grm}
R_{GRM}(x^i_{gen}) = f_{GRM}(x^i_{gen}) \odot f_{GRM}(x_{std}).
\end{equation}

\noindent To facilitate future research in reward-driven garment generation and fine-grained visual evaluation, we will publicly release the full Garment-RM model weights and training pipeline upon acceptance.

\begin{figure}[t]
  \centering
  \includegraphics[width=1\linewidth]{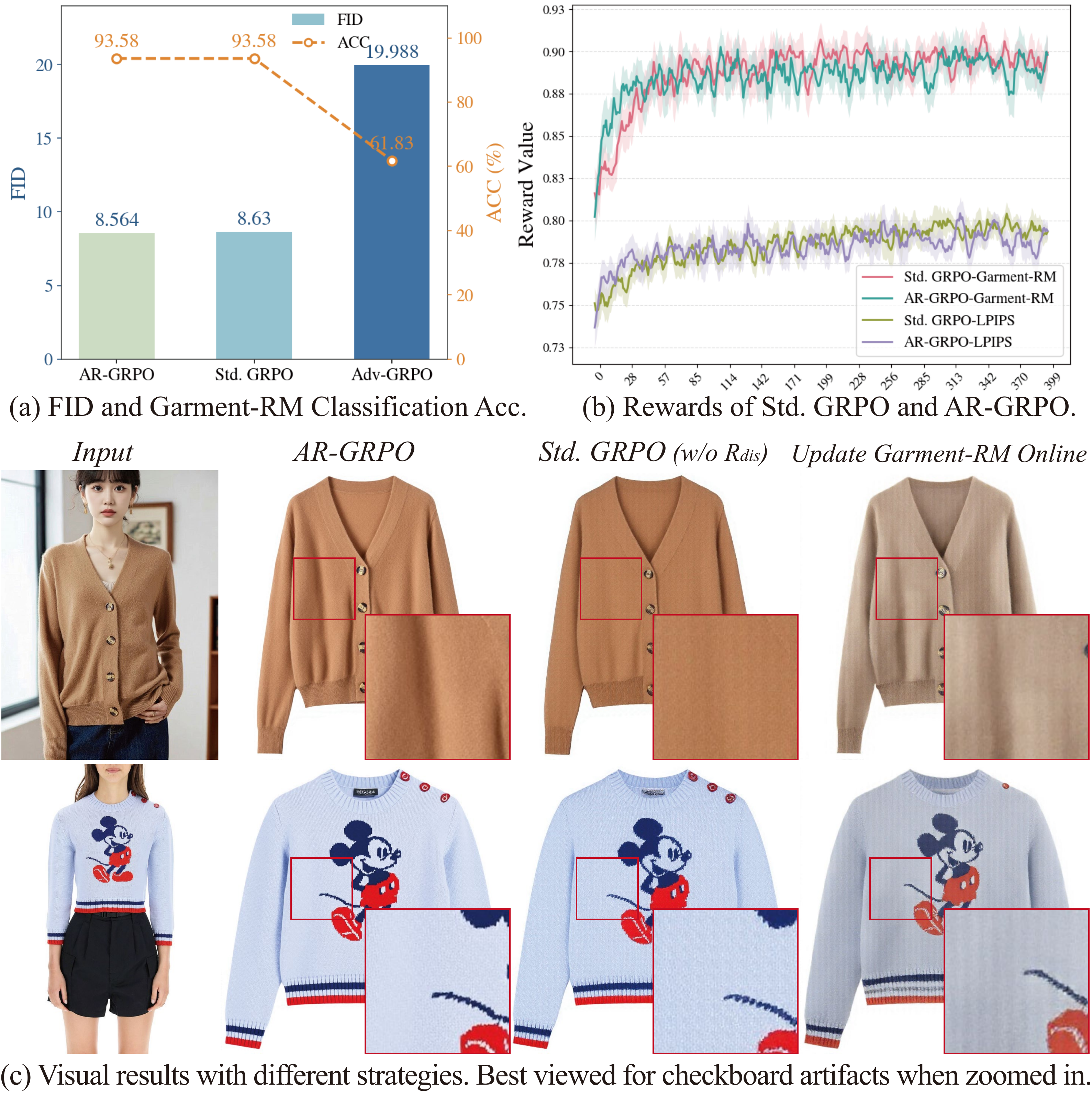}
  \caption{(a) Adv-GRPO~\cite{mao2025advgrpo} aggressively updates all reward models with adversarial loss, causing them to lose classification ability. (b) Reward trajectories: the discriminator first accelerates early reward rise, then stabilizes both $R_{GRM}$ and $R_{LPIPS}$ at an artifact-free equilibrium. (c) Standard GRPO and Adv-GRPO collapse into checkerboard artifacts, while AR-GRPO maintains clean outputs. Best viewed when zoomed in.}

  \label{fig: reward_hack}
\end{figure}

\subsection{Adversarial-Regularized GRPO (AR-GRPO)}
\label{sec: ar_grpo}

AR-GRPO is the core algorithmic contribution of this work, realizing the distribution-reshaping principle introduced in Sec.~\ref{sec:intro}. It fulfills the \textit{hacking prevention} prerequisite: ensuring that RL-driven exploration enriches only authentic high-frequency details without exploiting reward blind spots.

\noindent \textbf{From SFT Saturation to RL Post-Training.}
Although SFT establishes a strong initialization, we observe clear performance saturation: simply extending training or scaling data brings limited gains and may even degrade the generative prior due to noisy samples. \textit{(See Fig.~\ref{fig: motivation} for the qualitative comparisons post-SFT and the corresponding FID curves.)}. To further improve data efficiency and exploration ability, we adopt Group Relative Policy Optimization (GRPO) as a post-training strategy. Following prior RL works~\cite{liu2025flowgrpo, xue2025dancegrpo}, we introduce stochasticity through SDE-based sampling, which enables diversified trajectories and more effective policy optimization beyond deterministic SFT.

\noindent \textbf{Adversarial Supervision against Reward Hacking.}
Directly applying GRPO, however, leads to severe reward hacking: the reward score keeps increasing while visual quality deteriorates, often accompanied by checkerboard artifacts as in Fig.~\ref{fig: reward_hack}. To prevent the policy from exploiting imperfect reward signals, we introduce an adversarial discriminator following the spirit of Adv-GRPO~\cite{mao2025advgrpo}, with the pipeline in Fig.~\ref{fig: framework_grpo}. The discriminator provides an additional realism-aware reward that penalizes unnatural generations and is updated jointly with the diffusion policy, thereby stabilizing RL optimization and suppressing artifact collapse.

\noindent \textbf{Optimization Formulation.} 
We convert the deterministic ODE process lacking exploratory variance for GRPO, into an equivalent Stochastic Differential Equation (SDE)~\cite{liu2025flowgrpo}:
\begin{equation}
\begin{aligned}
\small
z_{t+\Delta t} = z_t + & \Biggl[ v_\theta(z_t, t, c)
+ \frac{\sigma_t^2}{2t}\Bigl(z_t + (1-t)v_\theta(z_t, t, c)\Bigr) \Biggr]\Delta t \\
\phantom{z_{t+\Delta t} = z_t} \quad & + \sigma_t \sqrt{|\Delta t|}\,\xi,
\end{aligned}
\label{eq:sde_sampling}
\end{equation}
where $\xi \sim \mathcal{N}(0, I)$ injects stochasticity for exploration, and $\sigma_t$ is the diffusion coefficient scheduling the noise level. Using this SDE formulation, we roll out $G=8$ trajectories for $T=10$ steps, starting from pure noise at $t=1$ down to $t=0$, to obtain the terminal latents $z_0^i$. These latents are subsequently decoded into generated images $x_{gen}^i = \mathcal{D}(z_0^i)$.

We formulate a comprehensive multi-reward system comprising two static perceptual rewards and a dynamically updated adversarial reward. As defined in Eq.~\ref{eq: reward_grm}, the static reward $R_{GRM}$ enforces detail fidelity via feature cosine similarity. As demonstrated in Fig.~\ref{fig: interpolation_top1_accuracy}, the combination of LPIPS~\cite{zhang2018perceptual} and Garment-RM yields sample selections that optimally align with human preferences. Furthermore, diversifying the optimization targets via a multi-reward formulation inherently increases the difficulty of reward hacking. Consequently, we incorporate a auxiliary perceptual reward $R_{LPIPS} = 1 - \text{LPIPS}(x^i_{gen}, x_{std})$, assigning it an equal weight to $R_{GRM}$. Conversely, the dynamic reward $R_{dis} = h_\phi(x^i_{gen})$, derived from the scalar output of our adversarial discriminator, provides critical authenticity supervision to systematically prevent artifact hacking. To stabilize optimization, each raw reward $* \in \{GRM, LPIPS, dis\}$ is independently normalized within the sampled group to yield its corresponding advantage:
\begin{equation}
\small
\hat{A}_{*}^{i} = \frac{R_{*}(x^i_{gen}) - \text{mean}(\{R_{*}(x^j_{gen})\}_{j=1}^G)}{\text{std}(\{R_{*}(x^j_{gen})\}_{j=1}^G)}.
\label{eq:advantage}
\end{equation}

The generative policy (diffusion model) is then updated via the clipped policy gradient objective:
\begin{equation}
\begin{aligned}
\mathcal{J}_{gen}(\theta) = \mathbb{E} &\Bigg[ \frac{1}{G} \sum_{i=1}^G \frac{1}{T} \sum_{t=1}^{T}
\min\Big(r_t^i(\theta) \hat{A}_{mix}^i, \\
&\text{clip}(r_t^i(\theta), 1-\epsilon_{clip}, 1+\epsilon_{clip}) \hat{A}_{mix}^i\Big) \Bigg],
\end{aligned}
\label{eq:grpo_obj}
\end{equation}
where the clipping threshold $\epsilon_{clip}$ explicitly bounds the policy updates to ensure stable optimization. In this objective, the combined advantage $\hat{A}_{mix}^i$ is computed as the weighted sum of the three independently normalized advantages:
\begin{equation}
\hat{A}_{mix}^i = \lambda_1 \hat{A}_{GRM}^i + \lambda_2 \hat{A}_{LPIPS}^i + \lambda_3 \hat{A}_{dis}^i.
\label{eq: adv_mix}
\end{equation}
Additionally, $r_t^i(\theta)$ denotes the importance ratio, formulated as the transition probability quotient between the current and the old generative policy during the denoising process:
\begin{equation}
r_t^i(\theta) = \frac{p_\theta(z_{t-\Delta t}^i | z_t^i, c)}{p_{\theta_{old}}(z_{t-\Delta t}^i | z_t^i, c)}.
\label{eq:importance_ratio}
\end{equation}

\noindent \textbf{Adversarial Discriminator Design and Update.} 
To construct the adversarial discriminator $h_\phi(\cdot)$, we adopt a frozen pre-trained DINOv2~\cite{oquab2023dinov2} backbone attached to a lightweight binary classification head comprising two fully connected layers. As demonstrated in Tab.~\ref{tab: human_preference_acc}, the DINO \texttt{[CLS]} token exhibits a superior capacity for filtering bad cases; therefore, we discard the patch tokens entirely. For the discrimination objective, we employ a hinge loss, which provides uniform and stable gradients. Furthermore, we intentionally abandon the Sigmoid activation, ensuring that the discrimination scores remain distinguishable even in high-confidence regions, thereby providing meaningful relative superiority judgments within the generated group. 

The discriminator $h_\phi(\cdot)$ is continuously updated using the following adversarial hinge loss:
\begin{equation}
\begin{aligned}
\small
\mathcal{L}_{disc}(\phi) =\;& \mathbb{E}_{x_{pos}} \big[\max(0, 1 - h_\phi(x_{pos}))\big] \\
&+ \mathbb{E}_{x_{gen}} \big[\max(0, 1 + h_\phi(x_{gen}))\big].
\end{aligned}
\label{eq:disc_loss}
\end{equation}
In practice, we first initiate the discriminator offline for 3000-step training with pre-saved $x_{pos}$ and SDE rollouts $x_{gen}$. During AR-GRPO, we update the discriminator for three steps before each generator update, so that the realism reward remains sufficiently strong and up-to-date during RL training. The positive samples $x_{pos}$ are drawn from a mixture of real natural images and the base model's full 25-step ODE inference results, while the negative samples are taken from the current 10-step SDE rollouts $x_{gen}$. This dynamic co-evolution constrains the generator's exploration within the manifold of authentic images.

\begin{algorithm}[t]
\caption{Adversarial-Regularized GRPO (AR-GRPO)}
\label{alg:ar_grpo}
\small
\begin{algorithmic}[1]
\Require DiT Policy $\theta$ (ref: $\theta_{old}$, untrained frozen base: $\theta_{base}$), discriminator $\phi$, Garment-RM, LPIPS.
\Require Params: Group size $G=8$, SDE steps $T=10$, generator update ratio $R=4$, weights $\lambda_{1:3}$, $\epsilon_{clip}$.
\State Initialize iteration counter $k=0$.
\Statex \hspace{-\algorithmicindent} {\textbf{While} not converged}

    \Statex \textbf{Phase 1: SDE \& ODE Rollout Sampling}
    \State Sample condition $c$ and ground-truth $x_{std}$.
    \State Generate $x_{base}$ via 25-step ODE using model $\theta_{base}$.
    \State Sample $z_1^i \sim \mathcal{N}(0, I)$ for $i \in [1, G]$.
    \State Solve SDE (Eq.~\ref{eq:sde_sampling}) to get $z_0^i$, decode $x_{gen}^i = \mathcal{D}(z_0^i)$.

    \State Increment counter: $k \leftarrow k + 1$.

    \If{$k \bmod R \neq 0$}
        \Statex \textbf{Phase 2: Discriminator Training}
        \State Form batches: $x_{pos} \!=\! \{x_{std}, x_{base}\}$, $x_{neg} \!=\! \{x_{gen}^i\}_{i=1}^G$.
        \State Update $\phi$ by minimizing $\mathcal{L}_{disc}(\phi)$ (Eq.~\ref{eq:disc_loss}).
    \Else
        \Statex \textbf{Phase 3: Multi-Reward Advantage Computation}
        \State Compute $R_{GRM}^i$, $R_{LPIPS}^i$, and $R_{dis}^i = h_\phi(x_{gen}^i)$.
        \State For $* \in \{GRM, LPIPS, dis\}$, normalize with mean $\mu_*$ and std $\sigma_*$:
        \Statex \hspace{\algorithmicindent} $\hat{A}_{*}^{i} = (R_{*}^i - \mu_{*}) / (\sigma_{*} + 1e^{-5})$.
        
        \Statex \textbf{Phase 4: Optimization \& Model Update}
        \State Compute combined advantage:
        \Statex \hspace{\algorithmicindent} $\hat{A}_{mix}^i = \lambda_1 \hat{A}_{GRM}^i + \lambda_2 \hat{A}_{LPIPS}^i + \lambda_3 \hat{A}_{dis}^i$.
        \State Compute ratio:
        \Statex \hspace{\algorithmicindent} $r_t^i(\theta) = p_\theta(z_{t-\Delta t}^i \mid z_t^i, c) / p_{\theta_{old}}(z_{t-\Delta t}^i \mid z_t^i, c)$.
        \State Update $\theta$ by maximizing the clipped surrogate objective:
        \Statex \hspace{\algorithmicindent} \small $\mathcal{J}_{gen} \!=\! \sum_{i,t} \min\big(r_t^i \hat{A}_{mix}^i, \text{clip}(r_t^i, 1\!-\!\epsilon_{clip}, 1\!+\!\epsilon_{clip}) \hat{A}_{mix}^i\big)$.
        \State Sync reference policy: $\theta_{old} \leftarrow \theta$.
    \EndIf

\Statex \hspace{-\algorithmicindent} {\textbf{end while}}
\end{algorithmic}
\end{algorithm}

\section{Experiments}

\begin{figure*}[t]
  \centering
  \includegraphics[width=1\linewidth]{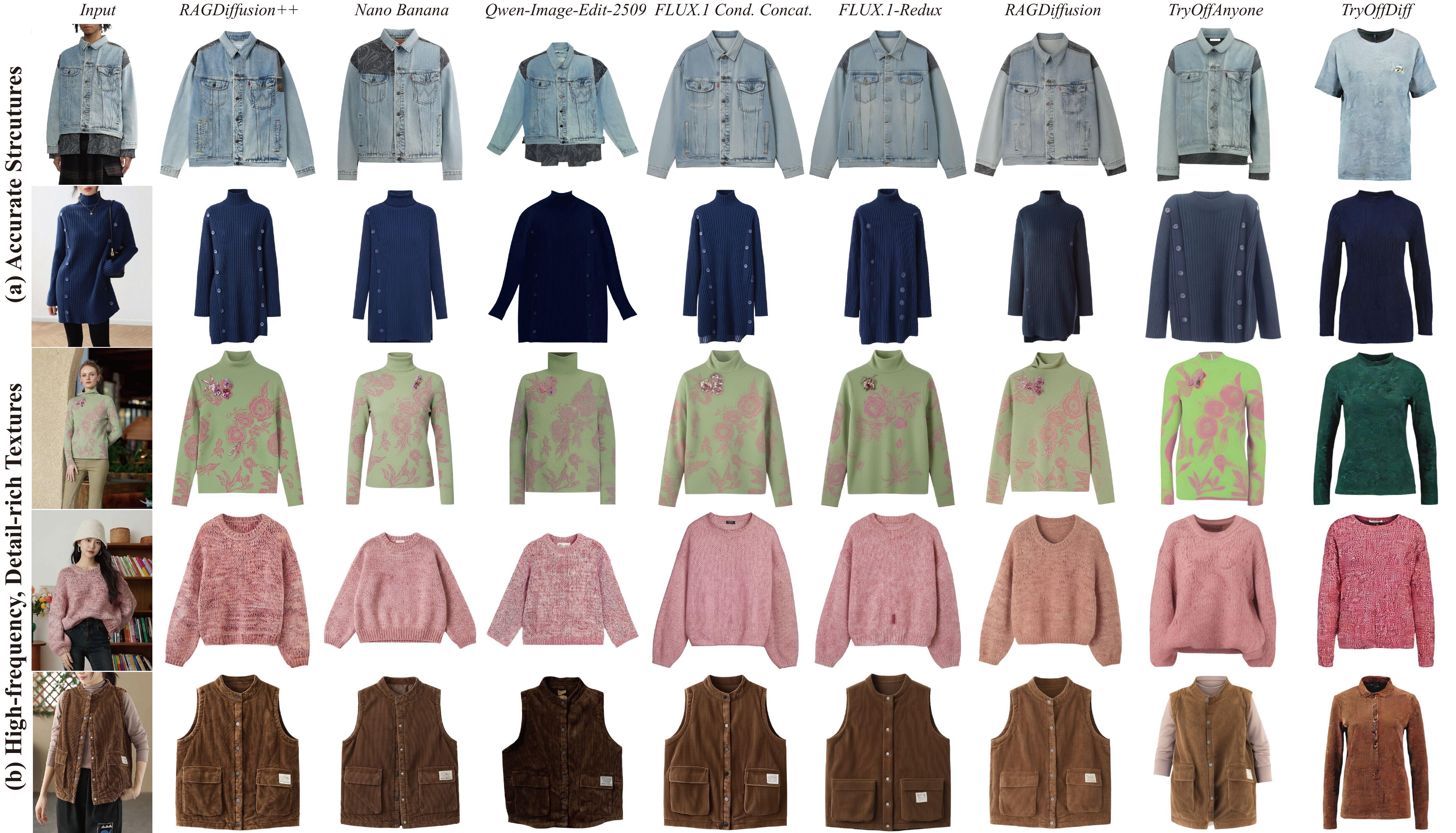}
  \caption{Qualitative comparison with state-of-the-art garment generation baselines. Best viewed when zoomed in.}
  \label{fig: main}
\end{figure*}

\begin{table}[t]
\centering
\fontsize{7pt}{8pt}\selectfont
\setlength{\tabcolsep}{6pt}  
\caption{Quantitative comparisons of generative performance against state-of-the-art baselines. The upper section evaluates SDXL-based models, while the lower section compares RAGDiffusion++ against FLUX-based and large-scale commercial models.} 
\begin{tabular}{lcccccc}
\toprule
\multicolumn{2}{l}{Method}       & SSIM $\uparrow$ & LPIPS $\downarrow$ & FID $\downarrow$ & KID $\downarrow$  & DISTS $\downarrow$ \\ 
\midrule
\multicolumn{7}{c}{\textit{SDXL Backbone (Trained on STGarment)}} \\
\midrule
\multicolumn{2}{l}{PBE}           & 0.5278 & 0.4821  & 26.55  & 10.08 & 0.352  \\
\multicolumn{2}{l}{ControlNet}    & 0.5225 & 0.4792  & 23.02  & 9.231 & 0.345  \\
\multicolumn{2}{l}{IP-Adapter}    & 0.6067 & 0.4728  & 14.50  & 3.747 & 0.264  \\
\multicolumn{2}{l}{TryOffDiff}    & 0.6124 & 0.5007  & 34.34  & 14.19 & 0.335  \\
\multicolumn{2}{l}{TryOffAnyone}  & 0.6575 & 0.4335  & 18.86  & 7.919 & 0.245  \\
\multicolumn{2}{l}{ReferenceNet}  & 0.6546 & 0.3979  & 10.70  & 1.399 & 0.224  \\
\multicolumn{2}{l}{RAGDiffusion}  & \textbf{0.6963} & 0.3684  & 9.990  & 1.092 & 0.192  \\
\midrule
\multicolumn{7}{c}{\textit{FLUX Backbone \& Large-Scale Commercial Models}} \\
\midrule
\multicolumn{2}{l}{Nano Banana 1} & 0.6576 & 0.3969 & 9.616 & 1.218 & 0.195 \\
\multicolumn{2}{l}{Qwen-Image-Edit}      & 0.6432 & 0.3893 & 10.55 & 1.466 & 0.203 \\
\multicolumn{2}{l}{FLUX.1-Redux}         & \underline{0.6717} & 0.4049 & 11.96 & 2.246 & 0.212 \\
\multicolumn{2}{l}{FLUX.1 Cond. Concat.} & 0.6684 & \underline{0.3605} & \underline{9.271} & \underline{1.036} & \underline{0.178} \\
\multicolumn{2}{l}{\textbf{RAGDiffusion++}} & 0.6688 & \textbf{0.3484} & \textbf{8.564} & \textbf{0.719} & \textbf{0.160} \\
\bottomrule
\end{tabular}
\label{tab: main}
\end{table}

\subsection{Experimental Setup}
\label{sec: setup}

\noindent \textbf{Datasets \& Implementation Details.} 
We train and evaluate RAGDiffusion++ on \textbf{STGarment-Plus}, which contains 25,756 training pairs of in-the-wild upper-garment images and their corresponding standard flat-lay targets, along with 1,969 test pairs (data distribution shown in Fig.~\ref{fig: dataset_comare}). The model is initialized from the pre-trained FLUX.1-dev DiT and trained at a $768\times768$ resolution using AdamW ($lr=5\times10^{-5}$) with a cosine scheduler. During the 10-day SFT phase on 16 NVIDIA H20 GPUs equipped with DeepSpeed ZeRO-2~\cite{deepspeed}, the sampling ratio of STGarment-Plus is linearly increased from 50\% until it fully replaces the STGarment mixture. For the subsequent AR-GRPO stage, we sample $G=8$ trajectories of $T=10$ steps per prompt (9 SDE steps, 1 ODE step) at data batch size of $bs=4$. The discriminator ($lr=1\times10^{-4}$) and generator ($lr=3\times10^{-6}$) are updated with a 3:1 step ratio over 2-3 days on 8 NVIDIA H20 GPUs. In Eq.~\ref{eq: adv_mix}, we set $\lambda_1=\lambda_2=\lambda_3=1$.  During DiT update, the gradient accumulation step is set to $bs \times G \times (T-1)=288$, such that one optimizer update is performed using the gradients accumulated from all sampled trajectories associated with a single data batch. As for testing, we utilize the Flow Match Euler Discrete Scheduler~\cite{flux} for 25 steps. Additionally, our Garment-RM is trained on 500,000 consensus-annotated images using an EfficientNet-style backbone on 4 GPUs, converging in approximately one day. All classification and retrieval losses are equally weighted during training.

\noindent \textbf{Baselines.} 
We comprehensively benchmark our methods across both generation quality and reward accuracy. For generation, our SDXL-based RAGDiffusion is compared against the official checkpoints of TryOffDiff~\cite{velioglu2024tryoffdiff} and TryOffAnyone~\cite{xarchakos2024tryoffanyone}, alongside four classic conditional methods (IP-Adapter~\cite{ye2023ipadapter}, ReferenceNet~\cite{hu2023animate}, ControlNet~\cite{zhang2023controlnet}, and PBE~\cite{yang2023paintbyexample}) retrained on our dataset. For the FLUX backbone, RAGDiffusion++ is evaluated against FLUX.1-Redux~\cite{flux-redux}, a CAT-VTON-inspired~\cite{chong2024catvton} token concatenation variant FLUX.1 Cond. Concat. (both with retrieved silhouette mask $\hat{L}_{sil}$), and zero-shot large-scale commercial models (Nano Banana 1~\cite{banana} and Qwen-Image-Edit~\cite{wu2025qwenimagetechnicalreport}). Notably, these massive commercial models inherently benefit from orders of magnitude more training data and compute. For the human preference prediction task, Garment-RM is benchmarked against generic vision foundation models (CLIP~\cite{radford2021clip}, DINO~\cite{simeoni2025dinov3}) and established aesthetic reward models (HPS v2.1~\cite{wu2023hpsv2}, ImageReward~\cite{xu2023imagereward}, and PickScore~\cite{kirstain2023pick}).

\noindent \textbf{Evaluation Protocols.} 
To assess generation quality, we utilize LPIPS~\cite{zhang2018perceptual} and SSIM~\cite{wang2004ssim} for reconstruction accuracy, DISTS~\cite{dists} for perceptual and structural similarity, and FID~\cite{parmar2021cleanfid} / KID~\cite{sutherland2018kid} for distributional realism. To evaluate the reward model, we measure its alignment with human judgments via Preference Accuracy, and employ Recall@$K$ and Filter@$K$ metrics to explicitly quantify its capability to retrieve high-quality garment generations and filter low-quality generations from diverse candidate pools.

\begin{table*}[t]
\centering
\caption{Quantitative ablation study of network components, architecture and data upgrades, and AR-GRPO rewards.} 
\begin{tabular}{lcccccc}
\toprule
\multicolumn{2}{l}{Method}          & SSIM $\uparrow$ & LPIPS $\downarrow$ & FID $\downarrow$ & KID $\downarrow$  & DISTS $\downarrow$ \\ 
\midrule
\multicolumn{7}{c}{\textit{Base Model Components in RAGDiffusion}} \\
\midrule
\multicolumn{2}{l}{Vanilla Model}        & 0.6576 & 0.3961 & 10.65 & 1.405 & 0.222 \\
\multicolumn{2}{l}{+ emb}                & 0.6614 & 0.3917 & 10.55 & 1.354 & 0.217 \\
\multicolumn{2}{l}{+ emb + sil}          & 0.6926 & 0.3690 & 10.22 & 0.940 & 0.210 \\ 
\multicolumn{2}{l}{RAGDiffusion}   & 0.6963 & 0.3684 & 9.990 & 1.092 & 0.192 \\ 
\midrule
\multicolumn{7}{c}{\textit{Data and Architecture Upgrades (SFT Phase)}} \\
\midrule
\multicolumn{2}{l}{+ FLUX Backbone}      & \underline{0.6718}      & 0.3613      & 9.795     & 1.122     & 0.170     \\
\multicolumn{2}{l}{+ STGarment-Plus Data}& 0.6586      & 0.3598      & 9.389     & 1.003     & 0.165     \\
\midrule
\multicolumn{7}{c}{\textit{AR-GRPO Optimization Ablations (RL Phase)}} \\
\midrule
\multicolumn{2}{l}{AR-GRPO w/ $R_{DINO}$}               & 0.6486 & 0.3829 & 12.18 & 2.731 & 0.202 \\
\multicolumn{2}{l}{AR-GRPO w/o $R_{GRM}$}               & 0.6650 & \underline{0.3509} & 8.989 & 0.904 & 0.163 \\
\multicolumn{2}{l}{AR-GRPO w/o $R_{dis}$ (Std. GRPO)}   & \textbf{0.6721} & 0.3561 & \underline{8.630} & \underline{0.769} & \underline{0.161} \\
\multicolumn{2}{l}{\textbf{RAGDiffusion++ (Full)}}      & 0.6688 & \textbf{0.3484} & \textbf{8.564} & \textbf{0.719} & \textbf{0.160} \\
\bottomrule
\end{tabular}
\label{tab: ablation}
\end{table*}

\begin{table}[t]
\centering
\fontsize{7pt}{8pt}\selectfont
\setlength{\tabcolsep}{6pt}  
\caption{Cross-dataset evaluation on Viton-HD and DressCode (DC). Models marked with * denote inner-dataset testing (trained on the target dataset), while others are evaluated in a cross-dataset setting.}
\begin{tabular}{l c c c c c c c}
\toprule
\multicolumn{2}{l}{Dataset} &\multicolumn{2}{c}{Viton-HD} &\multicolumn{2}{c}{DC-Upper} &\multicolumn{2}{c}{DC-Dress} \\
\cmidrule(lr){3-4} \cmidrule(lr){5-6} \cmidrule(lr){7-8} 
\multicolumn{2}{l}{Method} & FID $\downarrow$ & KID $\downarrow$ & FID $\downarrow$ & KID $\downarrow$ & FID $\downarrow$ & KID $\downarrow$ \\
\midrule
\multicolumn{8}{c}{\textit{SDXL-based Models}} \\
\midrule
\multicolumn{2}{l}{PBE}           & 80.8 & 52.9 & 45.0 & 21.1 & 105.1 & 74.6 \\  
\multicolumn{2}{l}{ControlNet}    & 69.9 & 43.2 & 48.9 & 23.6 & 118.3 & 95.4 \\  
\multicolumn{2}{l}{IP-Adapter}    & 46.8 & 17.8 & 28.8 & 11.1 & 47.8 & 19.0 \\  
\multicolumn{2}{l}{TryOffDiff*}    & \textcolor{gray}{14.0} & \textcolor{gray}{3.31} & 38.9 & 16.7 & 74.3 & 48.6 \\  
\multicolumn{2}{l}{TryOffAnyone*}  & \textcolor{gray}{\textbf{11.4}} & \textcolor{gray}{1.97} & 24.4 & 9.30 & 38.4 & 20.4 \\  
\multicolumn{2}{l}{ReferenceNet}  & 15.2 & 3.23 & \underline{17.8} & 6.24 & 51.2 & 22.4 \\
\multicolumn{2}{l}{RAGDiffusion}  & 12.3 & \underline{1.45} & \textbf{15.9} & \textbf{4.51} & 23.1 & 6.22 \\
\midrule
\multicolumn{8}{c}{\textit{FLUX Backbone \& Large-Scale Commercial Models}} \\
\midrule
\multicolumn{2}{l}{FLUX.1-Redux}      & 12.6 & 2.24 & 20.7 & 6.73 & \underline{20.7} & \underline{4.69} \\
\multicolumn{2}{l}{FLUX.1 Cond. Concat.}    & \underline{11.8} & 1.88 & 20.7 & 6.49 & 21.0 & 4.90 \\
\multicolumn{2}{l}{\textbf{RAGDiffusion++}} & \textbf{10.5} & \textbf{1.31} & 18.6 & \underline{5.13} & \textbf{20.4} & \textbf{4.50} \\
\bottomrule
\end{tabular}
\label{tab: cross_data}
\end{table}

\subsection{Main Generative Results}

\noindent \textbf{Qualitative results.}
Fig.~\ref{fig: main} presents a qualitative comparison on the STGarment-Plus dataset.
TryOffDiff~\cite{velioglu2024tryoffdiff} suffers severe structural collapses, and TryOffAnyone~\cite{xarchakos2024tryoffanyone} exhibits inaccurate structures and distorted colors. RAGDiffusion, FLUX.1 Cond. Concat., and FLUX.1-Redux~\cite{flux-redux} produce coherent shapes but consistently exhibit overly smooth textures, confirming the absence of high-frequency sampling trajectories. Qwen-Image-Edit~\cite{wu2025qwenimagetechnicalreport} often hallucinates garment structures. Nano Banana 1~\cite{banana} demonstrates comparable robustness but remains slightly inferior in local detail accuracy and material expressiveness. In contrast, RAGDiffusion++ stabilizes global topology via retrieved structural priors and optimizes high-frequency details via AR-GRPO, achieving unprecedented fidelity in fine-grained patterns while rendering fabrics with lifelike tactile realism.

\noindent \textbf{Quantitative results.}
As indicated in Tab.~\ref{tab: main}, we quantitatively evaluate the generation quality of various methods on the STGarment-Plus dataset, demonstrating that RAGDiffusion++ significantly outperforms all baseline approaches across most metrics. 
Notably, while Nano Banana exhibits visual quality second only to RAGDiffusion++, its quantitative scores are lower than those of the FLUX.1 Cond. Concat. baseline. We attribute this discrepancy to a "structure distribution gap" between Nano Banana~\cite{banana} training data and our test data. This is less of an issue for FLUX.1 Cond. Concat., which directly benefits from the retrieved information as described in Sec.~\ref{sec: preliminary}.
Furthermore, we observe that SSIM consistently fails to align with perceived visual quality. Its structure term is intrinsically sub-pixel-phase-sensitive, which paradoxically rewards both over-smooth outputs and regular artifact patterns (e.g., checkerboard grids) while penalizing realistic high-frequency textures (detailed analysis in the Appendix). We therefore treat SSIM as a coarse contour-fidelity proxy and rely on LPIPS, DISTS, FID, KID, and human evaluation (Sec.~\ref{sec:exp_rm}) as the primary indicators of micro-textural realism.

\noindent \textbf{Cross-dataset evaluation.}
Despite zero exposure to the Viton-HD~\cite{choi2021vitonhd} and DressCode~\cite{he2024dresscode} datasets during training, our framework exhibits strong out-of-distribution (OOD) generalization without any fine-tuning (Tab.~\ref{tab: cross_data}). We note that on DC-Upper, the SDXL-based RAGDiffusion attains marginally lower FID/KID---this is expected because DC-Upper is dominated by simple low-frequency cotton T-shirts, and AR-GRPO redistributes probability toward the high-frequency tail. On the more challenging Viton-HD and DC-Dress, RAGDiffusion++ recovers its advantage.

\begin{figure*}[t]
\centering
\begin{minipage}[t]{0.64\linewidth}
\centering
\includegraphics[width=1\linewidth]{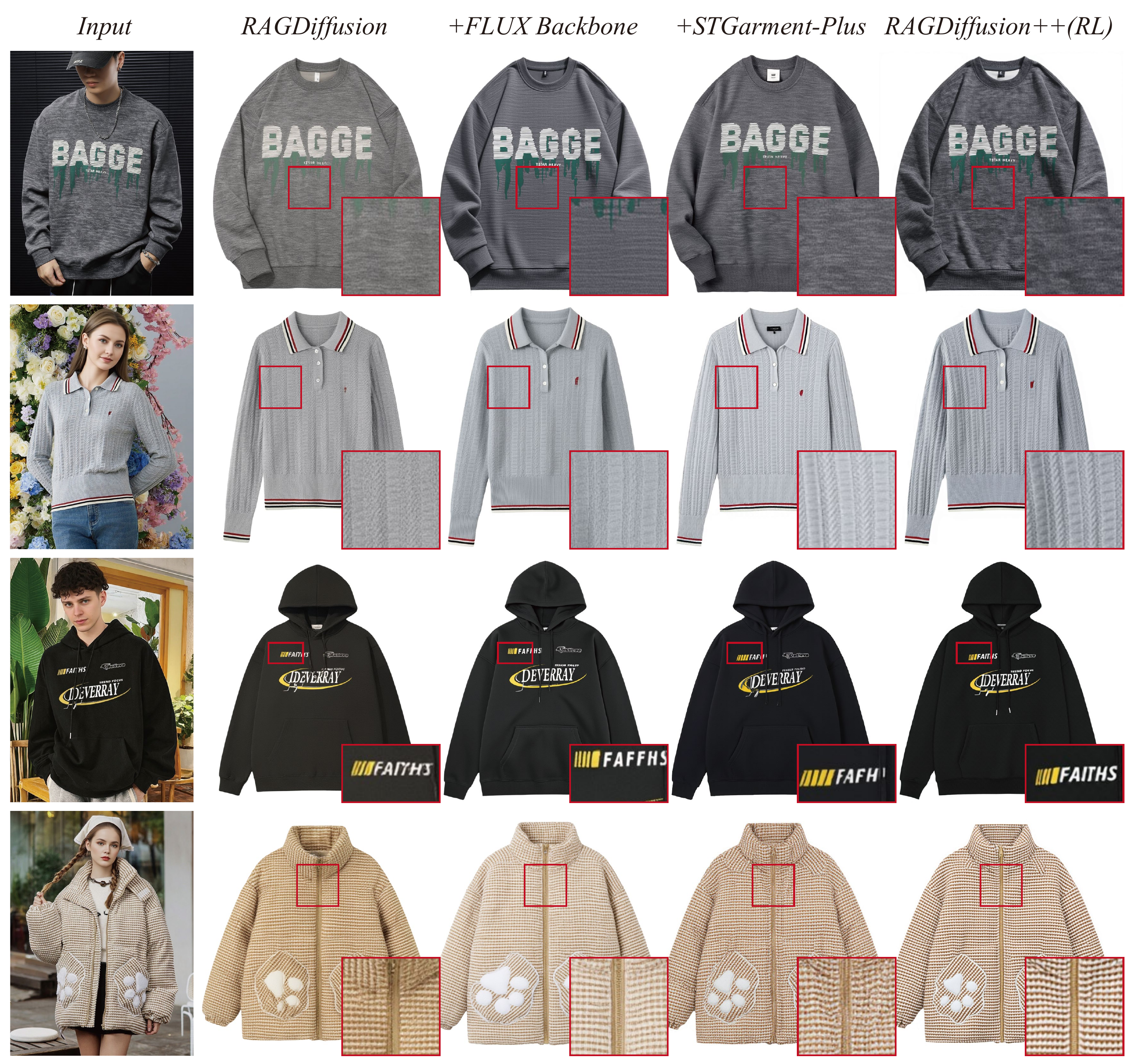}
\caption{The architecture update and data improvement during the SFT phase resulted in enhancements to richer lighting dynamics and fabric details. Best viewed when zoomed in.}
\label{fig: ablation_flux}
\end{minipage}
\begin{minipage}[t]{0.33\linewidth}
\centering
\includegraphics[width=\linewidth]{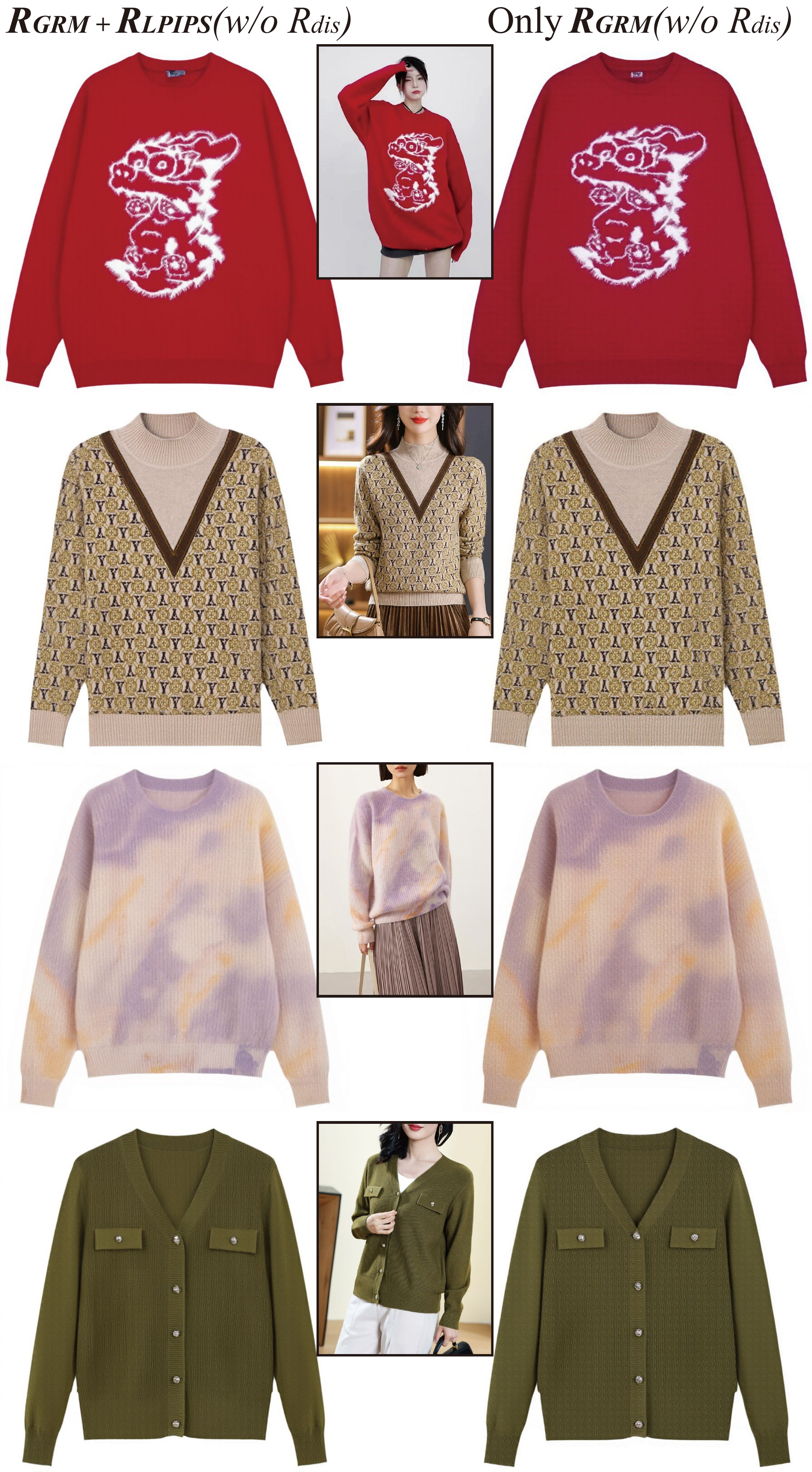}
\caption{We observe that $R_{GRM} + R_{LPIPS}$ delays the checkerboard artifacts.}
\label{fig: ablation_lpips}
\end{minipage}
\end{figure*}

\begin{figure*}[t]
  \centering
  \includegraphics[width=1\linewidth]{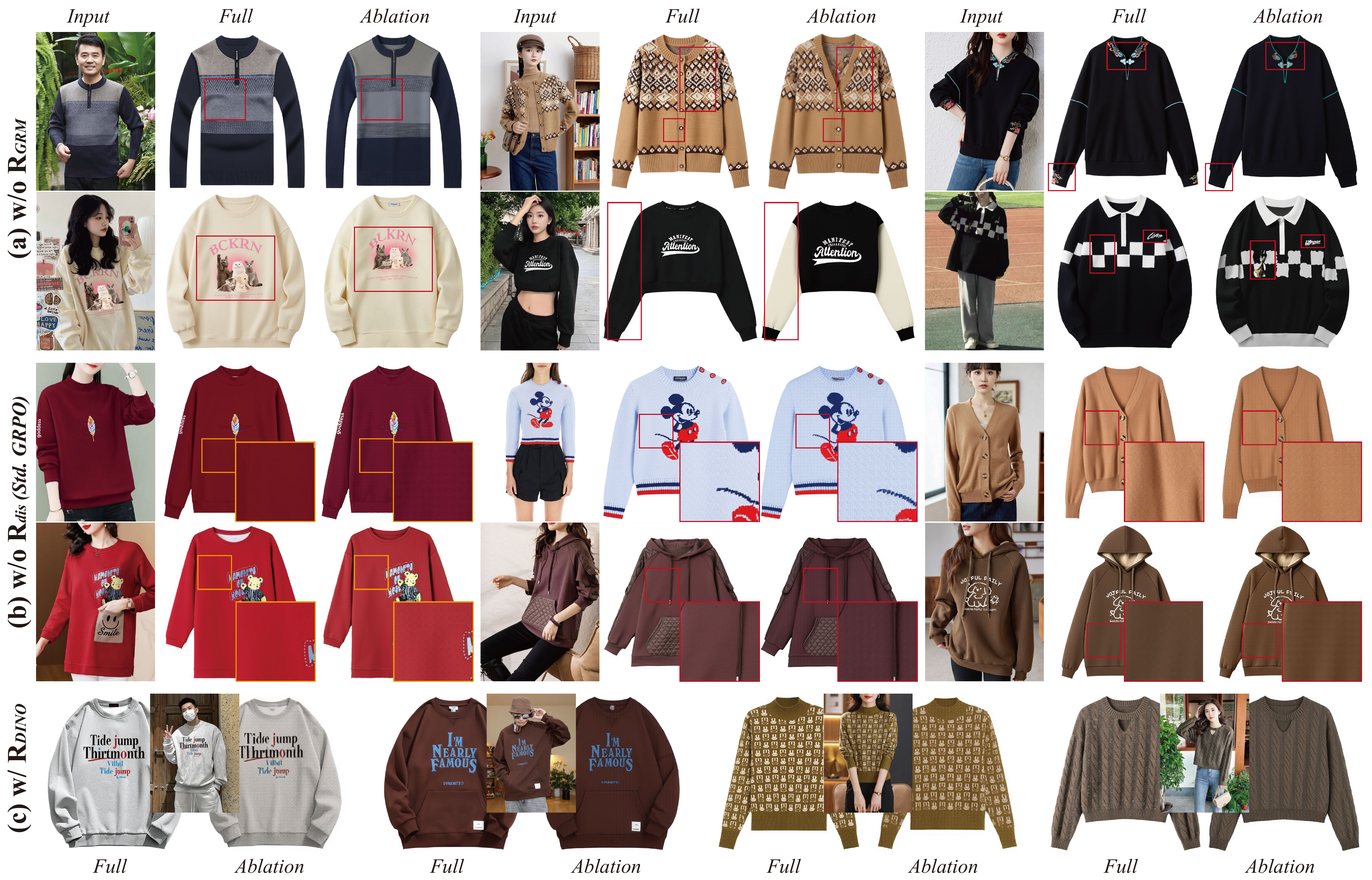}
  \caption{\textbf{Ablation of core reward components in AR-GRPO.} By comparing (a) the absence of Garment-RM (resulting in blurred textures and local misalignment), (b) the absence of the discriminator $R_{dis}$ (degrading to standard GRPO~\cite{liu2025flowgrpo} and causing dense checkerboard artifacts), and (c) the use of a generic DINO [CLS] reward (leading to severe trajectory distortion), we prove the indispensability of specialized reward models and adversarial regularization in preserving fabric realism and preventing reward hacking. Best viewed when zoomed in.}
  \label{fig: ablation_rl}
\end{figure*}

\subsection{Comprehensive Ablation Study}
\label{sec: ablation}

\noindent \textbf{Ablation of the Data Improvement and Architecture Update.}
We ablate the architectural and data quality evolution leading up to AR-GRPO in Fig.~\ref{fig: ablation_flux} and Tab.~\ref{tab: ablation}. Upgrading the base architecture from the legacy RAGDiffusion~\cite{li2025ragdiffusion} (SDXL) to the Dual-Image-Stream DiT yields images with significantly enhanced clarity and richer lighting dynamics; however, the fabrics still exhibit an unnatural, smooth plastic-like feel. The integration of the STGarment-Plus dataset further enriches the base fabric details within the FLUX backbone~\cite{flux}. Finally, the application of AR-GRPO pushes the generative quality to a higher ceiling. It dramatically enhances the fidelity of small logos and intricate patterns, while injecting unprecedented tactile realism into the textures, flawlessly restoring the physical appearance of complex woven and knitted materials.

\noindent \textbf{Ablation of AR-GRPO Reward Components.}
To validate the indispensable role of each component within our AR-GRPO framework, we design three variant settings (Fig.~\ref{fig: ablation_rl} and Tab.~\ref{tab: ablation}): 1) optimizing purely with a single pre-trained DINO [CLS] similarity reward (denoted as $w/ R_{DINO}$) as a intuitive solution; 2) removing our proposed Garment-RM score from RAGDiffusion++ ($w/o R_{GRM}$); and 3) removing the adversarial discriminator, which degrades the framework to standard GRPO ($w/o R_{dis}$).

Visual results indicate that utilizing a generic pre-trained DINO similarity as the sole reward is entirely counterproductive, leading to severe checkerboard artifacts and heavily distorted textures. Removing the Garment-RM ($w/o R_{GRM}$) yields overly smooth and blurred fabric details. Furthermore, it fails to accurately align fine-grained local structures like collars and cuffs, and triggers catastrophic visual collapses under strong lighting conditions. This proves that Garment-RM plays a decisive role in governing material quality, pattern fidelity, and sampling robustness. 
Critically, the absence of the adversarial discriminator ($w/o R_{dis}$) universally induces dense checkerboard artifacts (particularly visible when zoomed in on solid-color garments), a flaw that is devastating for industrial usability. These artifacts worsen progressively as training continues, heavily restricting the effective optimization window of standard GRPO~\cite{liu2025flowgrpo}. 

\noindent \textit{Why the gain of $R_{dis}$ appears small in Tab.~\ref{tab: ablation}.} The narrow FID/KID gap is expected: checkerboard noise produces near-Gaussian statistics in Inception feature space, and the regular grid inflates SSIM via the pathway discussed in the Appendix. The actual collapse is unambiguously visible in zoomed-in ablations (Fig.~\ref{fig: ablation_rl}(b) and Fig.~\ref{fig: reward_hack}), which we regard as the authoritative evidence for $R_{dis}$.

We further elucidate how the discriminator functions by visualizing the reward trajectories across DiT update steps in Fig.~\ref{fig: reward_hack}. Early in training, the pre-trained discriminator rapidly pushes the SDE sampling trajectory towards higher authenticity, accelerating the initial rise of both Garment-RM and LPIPS scores. In later stages, it serves as a strict authenticity constraint. By preventing the generator from exploiting the reward models (reward hacking), it stabilizes the scores at a slightly lower, yet physically accurate and artifact-free equilibrium.

\begin{figure*}[t]
  \centering
  \includegraphics[width=1\linewidth]{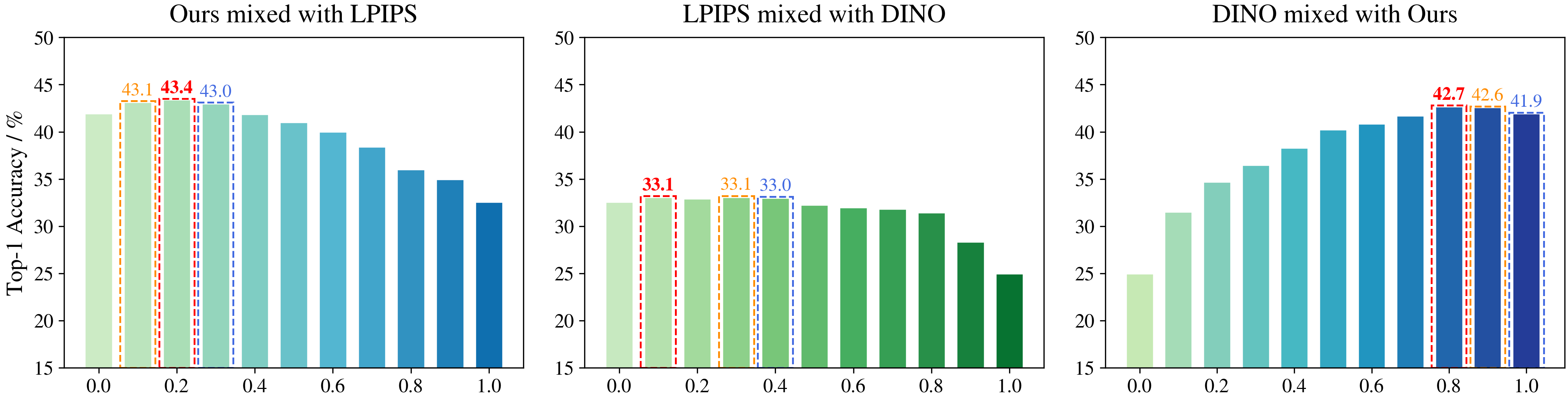}
  \caption{We observe that interpolation between different reward models can always get higher Top-1 accuracy. While Garment-RM achieves the best accuracy in a single reward setting, interpolating it with LPIPS score produces a bit better performance.}
  \label{fig: interpolation_top1_accuracy}
\end{figure*}

\noindent \textbf{Reward Auxiliary Objectives.}
Additionally, in Fig.~\ref{fig: ablation_lpips}, we justify the use of LPIPS as an auxiliary reward. In early middle-stage training without $R_{dis}$, the combination of $R_{GRM} + R_{LPIPS}$ mitigates and delays the onset of checkerboard artifacts compared to pure $R_{GRM}$, inherently increasing the difficulty for the model to ``cheat''. Furthermore, our interpolation analysis (Fig.~\ref{fig: interpolation_top1_accuracy}) proves that this combined reward correlates more robustly with sample quality.

\noindent \textbf{Ablation of Legacy RAG Components.}
As shown in Tab.~\ref{tab: ablation} (upper section), incrementally introducing structure embeddings and silhouette landmarks progressively improves contour accuracy and detail fidelity, confirming the complementary roles of these retrieval-based priors (detailed analysis in the Appendix). Furthermore, although RAGDiffusion is trained exclusively on upper-body data, incorporating lower-body/dress flat-lay images into the memory database enables zero-shot cross-category generalization (Tab.~\ref{tab: cross_data}), demonstrating the cost-effectiveness of the RAG paradigm.


\begin{figure*}[t]
  \centering
  \includegraphics[width=1\linewidth]{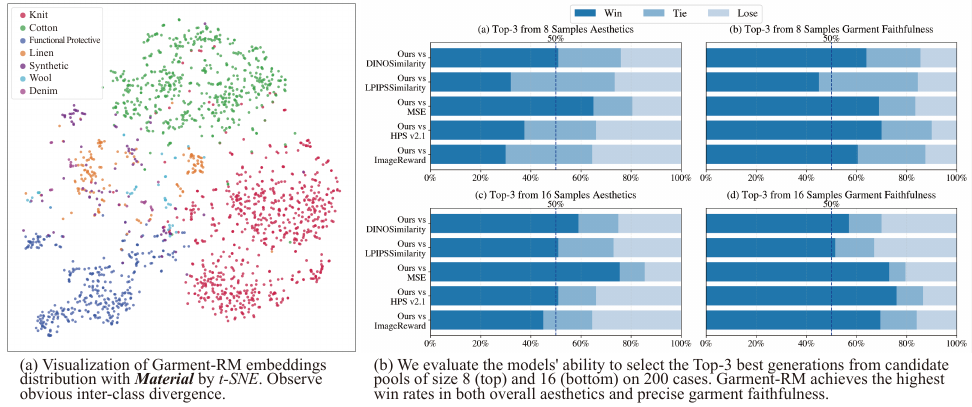}
  \caption{T-SNE figures of Garment-RM embeddings and human evaluation on preference selection to validate the effectiveness of Garment-RM.}
  \label{fig: merge_win_tistri}
\end{figure*}

\begin{table*}[t]
  \caption{Results of Garment-RM and comparison methods on human preference prediction. Preference accuracy and Recall/Filter scores are evaluated on a test set of 1,969 prompts (with 9 pairs per sample). All scores are averaged per prompt.}
  \label{tab: human_preference_acc}
  \centering
  \renewcommand\tabcolsep{9pt}
  \renewcommand\arraystretch{0.95}
  \small
  \begin{tabular}{l|c|ccc|ccc}
    \toprule
      \multirow{2}*{Model} & Preference & \multicolumn{3}{c|}{Recall} & \multicolumn{3}{c}{Filter} \\
      ~ & Acc. & @1 & @2 & @4 & @1 & @2 & @4 \\
    \midrule
      HPS v2.1 & 33.48 & 2.49 & 7.62 & 24.68 & 10.82 & 20.42 & 37.94 \\
      ImageReward & 38.95 & 4.27 & 10.97 & 29.91 & 14.27 & 23.36 & 42.46 \\
      PickScore & 45.04 & 6.91 & 15.79 & 39.00 & 19.86 & 31.59 & 49.16 \\
    \midrule
      MSE Score & 52.25 & 20.47 & 36.92 & 72.12 & 9.40 & 16.40 & 36.06 \\
      CLIP Similarity & 77.97 & 22.45 & 40.58 & 72.07 & 67.60 & 82.02 & 92.59 \\
      DINO (Patch) Sim. & 71.58 & 27.73 & 44.03 & 73.54 & 36.72 & 57.54 & 77.55 \\
      DINO (CLS) Sim. & \underline{81.22} & 24.94 & 44.74 & 77.55 & \textbf{70.85} & \textbf{86.39} & \textbf{95.94} \\
      LPIPS Similarity & 81.14 & \underline{32.55} & \underline{51.96} & \underline{82.88} & \underline{69.07} & 83.70 & 93.65 \\
    \midrule
      Garment-RM (Ours) & \textbf{84.67} & \textbf{41.90} & \textbf{57.59} & \textbf{84.92} & 68.66 & \underline{84.00} & \underline{94.26} \\
    \bottomrule
  \end{tabular}
\end{table*}

\begin{figure}[t]
  \centering
  \includegraphics[width=0.8\linewidth]{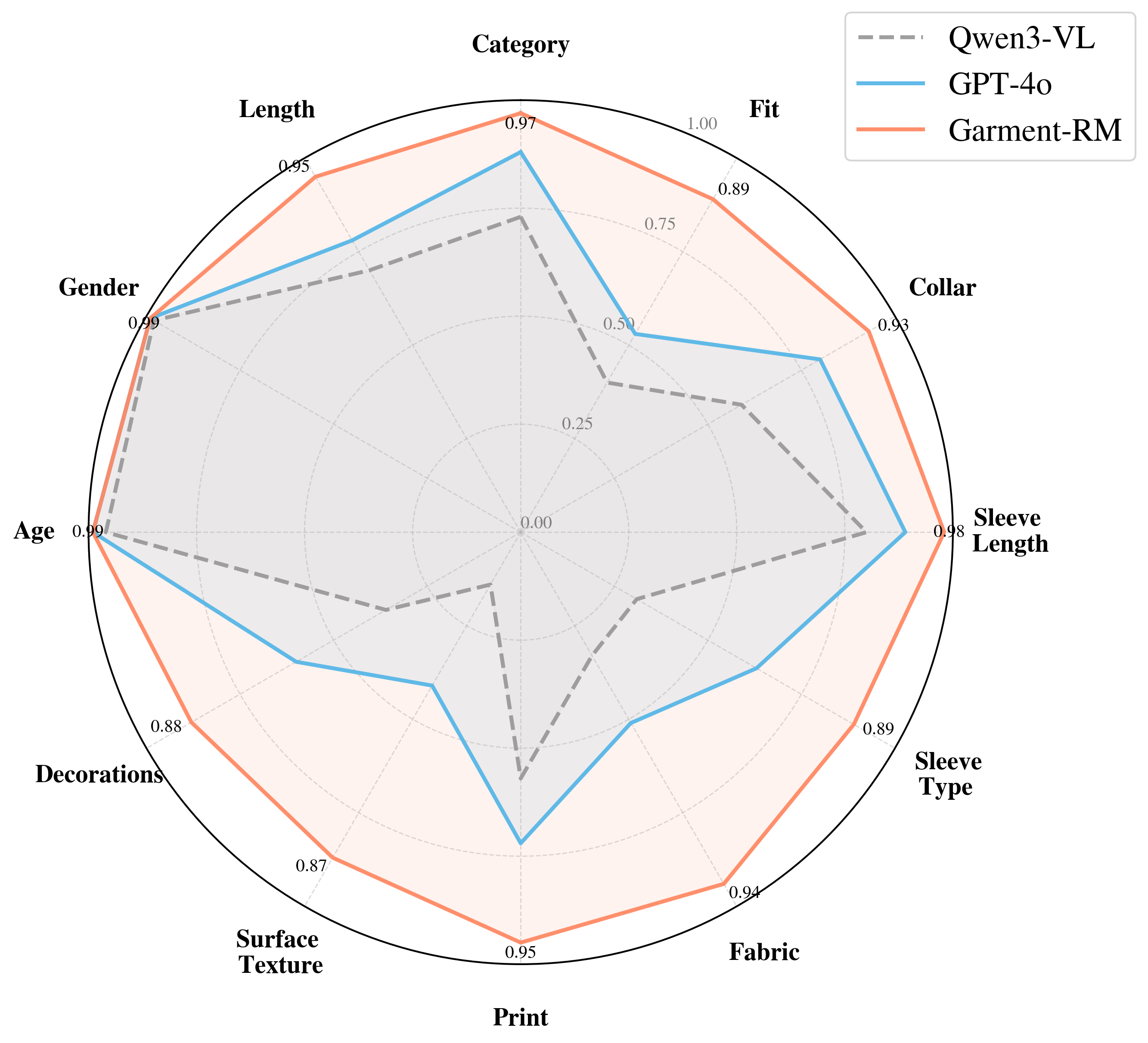}
  \caption{\textbf{Performance comparison of fine-grained attribute perception.} Garment-RM comprehensively outperforms state-of-the-art large language models (GPT-4o and Qwen3-VL), leading to reliable, attribute-aware signals for the RL optimization phase.}
  \label{fig: radar_accuracy}
\end{figure}

\subsection{Effectiveness of Garment-RM}
\label{sec:exp_rm}

To validate the effectiveness and necessity of our proposed Garment-RM, we conduct a comprehensive analysis comparing it against existing generic vision foundation models and aesthetic reward models.

\noindent \textbf{Attribute Classification Accuracy.} 
The superiority of Garment-RM stems from its multi-task consensus-driven recognition engine. In Fig.~\ref{fig: radar_accuracy}, we present a radar chart illustrating the classification accuracy across the 12 predefined fine-grained attributes (detailed in the Appendix). Compared to the large language models, Garment-RM covers a significantly larger area, achieving uniformly high accuracy across all fine-grained categories. This attribute-awareness ensures that the reward signal explicitly penalizes structural hallucinations or incorrect textures during the RL phase.

\noindent \textbf{Human Preference Accuracy.} 
We evaluate reward models on a test set of 1,969 samples (Tab.~\ref{tab: human_preference_acc}). For each sample, 8 candidate generations are collected using different models and noise perturbations; human annotators label the best and worst, from which we construct preference pairs and measure both preference accuracy and Recall@$K$/Filter@$K$ (detailed protocol in the Appendix). Widely used aesthetic reward models (HPS v2.1~\cite{wu2023hpsv2}, ImageReward~\cite{xu2023imagereward}, PickScore~\cite{kirstain2023pick}) achieve preference accuracy below 50\%, revealing a severe domain gap---they emphasize lighting and composition but fail to capture physical correctness and material fidelity. Generic vision models (CLIP~\cite{radford2021clip}, DINO~\cite{simeoni2025dinov3}) perform somewhat better yet remain inferior. In contrast, Garment-RM achieves the highest Preference Accuracy (84.67\%) and dominates the Recall metrics, demonstrating superior alignment with professional human judgments.

\noindent \textbf{Human Evaluation.} 
To deeply evaluate the ability of Garment-RM to select the more preferred images among generated images, we produce 8/16 candidate generations on 200 cases in the same way as Sec.~\emph{Human Preference Accuracy}, and use different methods to select from those images to get top3 results. Then three annotators rank these selected top-3 images. Fig.~\ref{fig: merge_win_tistri} (b) shows that Garment-RM achieves the highest
win rates, especially garment faithfulness, demonstrating that Garment-RM can select images that are more natural and high-fidelity. 

\noindent \textbf{Interpolation Analysis of reward combination.} 
When humans evaluate images, the selection process contains multiple elements such as fidelity, authenticity, etc. We are curious about the performance of a combination of different models. We test interpolation among the top3 models in Tab.~\ref{tab: human_preference_acc}, their accuracy on the test set is in Fig.~\ref{fig: interpolation_top1_accuracy}. We observe that interpolation between different reward models can always get higher Top-1 accuracy, and interpolating Garment-RM with LPIPS score reaches a bit better performance.

\noindent \textbf{Feature Space Discriminability.} 
We further visualize the learned Garment-RM embedding distribution in Fig.~\ref{fig: merge_win_tistri}(a) 
to gain an in-depth comprehension of how the embeddings help penalize incorrect textures. The features extracted by Garment-RM exhibit highly compact intra-class clustering and distinct inter-class separation. Such a well-structured feature space ensures that the cosine similarity reward $R_{GRM}$ (Eq.~\ref{eq: reward_grm}) provides a dense and meaningful gradient signal for the subsequent AR-GRPO optimization.

\subsection{Effectiveness of AR-GRPO}
\label{sec: effectiveness_ar_grpo}

\noindent \textbf{Breaking the SFT Saturation Bottleneck.}
To illustrate the necessity of transitioning from Supervised Fine-Tuning (SFT) to Reinforcement Learning (RL), we first analyze the inherent saturation bottleneck of the SFT paradigm by evaluating checkpoints from different SFT phases on a subtest set with 200 samples. As depicted in the training curves (Fig.~\ref{fig: motivation}), during the SFT phase, the generative model reaches a performance plateau, indicating that extending the training duration or scaling the data yields diminishing returns (increasing the data from STGarment may introduce low-quality samples). Qualitatively, models relying solely on SFT struggle to capture high-frequency physical details, often producing garments with blurred patterns, faded colors, and flattened, plastic-like textures. In stark contrast, the introduction of AR-GRPO shatters this ceiling, triggering a sharp decline in FID and a corresponding surge in reward (Fig.~\ref{fig: reward_hack}). The resulting RL-optimized images exhibit vibrant colors, razor-sharp typography, and highly faithful fabric textures, demonstrating that stochastic RL exploration is essential for mastering micro-level physical realism.

\noindent \textbf{Frequency-Domain Analysis.}
We further validate the improvements of AR-GRPO in the frequency domain by comparing the average Fourier spectra of the Base model (post-SFT with the FLUX backbone), the AR-GRPO model, and the Real test set. Specifically, each image is converted to grayscale, transformed via a 2D Fast Fourier Transform (FFT), and represented by the log-power spectrum:
\begin{equation}
\mathcal{F}(D)=\log\left(1+|\mathrm{FFT}(D)|^2\right).
\end{equation}
Averaging over all matched samples yields the dataset-level spectra $\mathcal{F}(D_{\mathrm{Base}})$, $\mathcal{F}(D_{\mathrm{RL}})$, and $\mathcal{F}(D_{\mathrm{Real}})$.

As shown in Fig.~\ref{fig: freq}(a), the Base model exhibits two clear issues. First, its error relative to real images is exceptionally large in the middle- and high-frequency regions, indicating a severe mismatch in texture and fine-detail statistics. Second, the error map contains obvious directional patterns rather than an isotropic distribution, suggesting a structured spectral bias. In contrast, the RL model produces significantly weaker residuals and less pronounced directional artifacts, confirming that AR-GRPO reduces both frequency-domain error and structural bias. To further quantify this effect, we compute the radially averaged spectral error curves in Fig.~\ref{fig: freq}(b). The RL model consistently achieves lower error over a broad range of middle and high frequencies, indicating a superior alignment with the real-image spectrum and mathematically explaining the enhanced tactile realism observed in the generated fabrics.

\section{Conclusion}
\label{sec:conclusion}
In this paper, we present RAGDiffusion++, an omni-level aligned framework that resolves the High-Frequency Trajectory Collapse in industrial-grade garment synthesis. Our core insight is that SFT converges to the conditional-mean mode dominated by smooth textures, making high-frequency patterns structurally un-sampleable; reinforcement learning breaks this collapse by reshaping the sampling distribution---stochastic SDE exploration combined with reward-weighted policy gradients elevates physically correct but low-probability trajectories into the sampleable regime. We operationalize this principle through three complementary prerequisites: (i)~a Dual-Image-Stream FLUX architecture trained on the challenging STGarment-Plus dataset to establish high-frequency priors (\textit{inherent capacity}); (ii)~Garment-RM, an attribute-aware reward model that provides fine-grained material-level optimization signals (\textit{perceptive reward}); and (iii)~AR-GRPO, which integrates an adversarial discriminator to detect and penalize reward exploitation (\textit{hacking prevention}).
Extensive experiments on STGarment-Plus, VITON-HD, and DressCode demonstrate that RAGDiffusion++ consistently achieves state-of-the-art performance, confirming the effectiveness of our distribution-reshaping paradigm. We will release the STGarment-Plus dataset and Garment-RM weights to foster future research in fine-grained generative alignment.

\bibliographystyle{IEEEtran}
\bibliography{main}

\end{document}